\documentclass{article}

\makeatletter
\def\input@path{{iclr2027/}}
\makeatother
\usepackage{iclr2027/iclr2027_conference,times}

\usepackage[T1]{fontenc}
\usepackage{amsmath}
\usepackage{amssymb}
\usepackage{array}
\usepackage{booktabs}
\usepackage{graphicx}
\usepackage{xcolor}
\definecolor{bestSky}{RGB}{184,225,250}
\newcommand{\best}[1]{%
  \begingroup\setlength{\fboxsep}{1pt}%
  \colorbox{bestSky}{#1}%
  \endgroup%
}
\usepackage{microtype}
\usepackage{natbib}
\usepackage{url}
\usepackage{hyperref}
\iclrfinalcopy
\usepackage{etoolbox}
\makeatletter
\patchcmd{\@maketitle}
  {Published as a conference paper at ICLR 2027}
  {Preprint}
  {}
  {\PackageError{arxiv_setup}{Could not replace the ICLR publication header}
    {Check the conference style before releasing this preprint.}}
\makeatother

\title{When Tools Silently Lie: Evaluating and Mitigating Blind Compliance in Tool-Augmented Data Agents}

\author{%
  Zifu Tao\thanks{This work was initiated while Zifu Tao was an undergraduate student at Tongji University.}\\
  Tongji University\\
  \href{mailto:zifutao@nyu.edu}{zifutao@nyu.edu}
  \And
  Changqing Yin\thanks{Corresponding author.}\\
  Tongji University\\
  \href{mailto:yinchangqing@tongji.edu.cn}{yinchangqing@tongji.edu.cn}
}

\newcommand{\researchoriginfootnote}{}

\date{}

\begin{document}

\maketitle
\researchoriginfootnote
\pagestyle{fancy}
\fancyhead{}
\fancyhead[L]{Preprint}
\thispagestyle{fancy}

\begin{abstract}
Tool-augmented data agents rely on tool outputs for analytical decisions. Yet successful execution can return plausible but incorrect evidence, requiring agents to decide whether to trust or verify it. Understanding this failure requires examining both the evidence obtained through checking and the answer ultimately adopted. We introduce ToxicBench to measure checking and adoption under numerical, label, schema, and retrieval errors, pairing clean and poisoned observations over fixed source data. In the 118-task GPT evaluation across three adapters, poisoning lowers task success by 26--39 percentage points. Ordinary retries help under one-shot poisoning, whereas repeated poisoning reveals wrong-answer adoption after checking. Controls on three public tables isolate how supplied evidence affects recovery. After freezing the scorer, we compare its judgments with human annotations on 200 trajectories, finding 96\% task-success agreement. Human judgments support retry gains over Base and confirm adoption after checking on audited tasks. We release trajectories, versioned scoring, and reference and delivery audits. These findings highlight evidence availability and answer selection as complementary dimensions of agent reliability.

\end{abstract}
\section{Introduction}
\label{sec:introduction}

Language-model agents increasingly delegate computation and interaction to tools \citep{schick2023toolformer,yao2023react,gao2023pal}. In data analysis, they inspect schemas, execute Python or SQL, and turn results into reports \citep{lei2025dacomp}. Delegating computation does not remove the need to judge evidence: a tool call can execute successfully while its returned observation supports an incorrect conclusion. The agent must decide not only which tool to call, but when its output is trustworthy enough to use.

Consider an agent asked which store had the highest revenue. The tool returns a plausible amount attached to the wrong store, although the source table still contains the correct store--value pair (Figure~\ref{fig:silent-tool-poisoning}). No exception or malformed response alerts the agent to the problem. Checking that the amount is positive or falls within a reasonable range will not resolve the incorrect binding; the agent needs evidence connecting the value to the right entity. Similar failures arise when an aggregate uses the wrong denominator or an apparently relevant data dictionary supplies an outdated column meaning. In each case, normal-looking tool output can turn into confident but incorrect analysis.

These failures motivate looking beyond outcomes to trajectories and error propagation \citep{mazaheri2026agentatlas,gurram2026agentprop}. A wrong final answer alone cannot distinguish accepting the first observation, checking against further bad evidence, or obtaining a correction but selecting the wrong conclusion. Conversely, counting additional tool calls does not establish recovery: an agent may repeat a computation, inspect unrelated rows, or consult the same unreliable source. Evaluation must track what was corrupted, what evidence became available afterward, and which answer the agent ultimately adopted.

Recent work studies silent tool errors, recoverable environment hazards, stage-aligned perturbations, and trajectory-level evidence use \citep{sun2024toolsfail,tian2026toolbenchx,zheng2026toolrobustbench,kim2025trace}. Our question builds on these directions: when a data agent receives corrupted evidence but retains access to the underlying data, does it check that evidence, and does the check change its final answer? A targeted benchmark must make these possibilities inspectable without conflating a corrupted tool response with a changed source table or assuming that any verification action supplies correct evidence.

We introduce \textbf{ToxicBench} to study this setting, which we call \emph{silent tool poisoning}. A proxy changes a successfully executed tool's returned observation while leaving the source data fixed. Each task pairs the same query, data, model, and adapter in clean and poisoned environments. Numerical interventions alter quantities, ratios, or rankings; semantic interventions alter entity bindings, column meanings, metadata, or retrieved evidence. Inspectable source tables and references let us relate the final answer to the intended corruption. Trajectory metrics then distinguish adoption without checking, adoption after a qualifying check, and recovery, rather than treating all incorrect answers or all additional calls alike.

Keeping the source fixed separates this diagnosis from source-data quality problems \citep{gu2025radar}: a wrong answer cannot simply reflect faithful analysis of altered tables. The clean answer remains well-defined and available through checking. Unlike indirect prompt injection \citep{greshake2023indirect,debenedetti2024agentdojo}, we alter returned evidence without asking the agent to follow malicious instructions; its analytical objective remains unchanged.

Feedback-based revision motivates asking agents to reconsider their answers \citep{shinn2023reflexion,madaan2023selfrefine,hamad2025toolcritic}, but verification requires a careful control. Under one-shot corruption, an ordinary retry can encounter clean evidence without a verification instruction. Attributing all improvement to checking would conflate the instruction with another opportunity to solve the task. We compare Double-pass, Verification-only, and Generic Guard under shared route counts, step limits, and per-request token caps, recording evidence along both routes. Repeated poisoning tests checks that can also be corrupted; experiments on three public tables separately vary supplied evidence. These comparisons examine when checking helps, not simply whether a checking prompt improves a score.

The experiments show why this distinction matters. Across LangGraph, smolagents, and AutoGen in the 118-task GPT evaluation, poisoning lowers task success by 26--39 percentage points despite strong clean performance. Cross-model evaluations extend the comparison to Claude and Qwen. An ordinary second attempt is a strong control under one-shot poisoning, while repeated poisoning reveals continued adoption of wrong answers after checking. A post-freeze human evaluation of 200 trajectories provides 96\% task-success scoring agreement and corroborates the retry benefit over Base on audited tasks and adoption after checking. Together, these results motivate evaluating the evidence a check returns alongside the decision that follows it.

Our contributions are:
\begin{itemize}
    \item An inspectable data-analysis benchmark that changes returned evidence while preserving source tables, including numerical, semantic/schema, and multi-table tasks.
    \item A distinction between checking and answer adoption: trace-level and human evidence show that a fresh check need not prevent adoption of a poisoned conclusion, especially when the check is also corrupted.
    \item Controls separating another attempt, verification instructions, and later evidence conditions, including controlled comparisons on three public tables that test how supplied evidence affects recovery.
\end{itemize}

\section{Related Work}
\label{sec:related}

Tool-use benchmarks evaluate API invocation \citep{qin2023toolllm,li2023apibank,patil2024gorilla}, while interactive and data-agent benchmarks assess task solving \citep{liu2023agentbench,zhou2023webarena,lei2025dacomp}. TRACE uses an evidence bank and LLM judges to assess trajectory efficiency, hallucination, and adaptivity \citep{kim2025trace}. ToxicBench shares this process perspective but distinguishes answers supported by returned observations from answers that are correct according to the source data, tracking checking and final adoption.

\emph{Tools Fail} studies silent errors \citep{sun2024toolsfail}; ToolBench-X and PALADIN address broader tool failures and recovery \citep{tian2026toolbenchx,vuddanti2025paladin}. ToolRobustBench provides stage-aligned, deterministic, cascade-aware diagnosis within a tool-calling episode \citep{zheng2026toolrobustbench}. Related threats target tool metadata \citep{li2026mcpitp,ye2026trustdesc} and retrieved content \citep{zou2024poisonedrag,liang2025saferag}. In financial agents, \citet{wu2026agentdrift} find that self-verification or contamination detection need not improve recommendations under manipulated tool data. ToxicBench studies related failures through paired data-analysis tasks, fixed source tables, and separate checking, adoption, and recovery metrics under one-shot and repeated corruption (Table~\ref{tab:closest-work}).

Data validation addresses source quality \citep{schelter2018automating,breck2019data,gu2025radar}; ToxicBench preserves sources as checking paths. Feedback and self-monitoring revise answers or tool use \citep{shinn2023reflexion,madaan2023selfrefine,zhou2024metarag,hamad2025toolcritic}, while calibration links confidence to tool decisions \citep{xuan2026confidence,subramani2025mice}. Our comparisons use the same budget limits to distinguish ordinary retries from explicit verification under different evidence conditions.

\section{ToxicBench}
\label{sec:benchmark}
\label{sec:problem}

\begin{figure}[t]
    \centering
    \includegraphics[width=\linewidth]{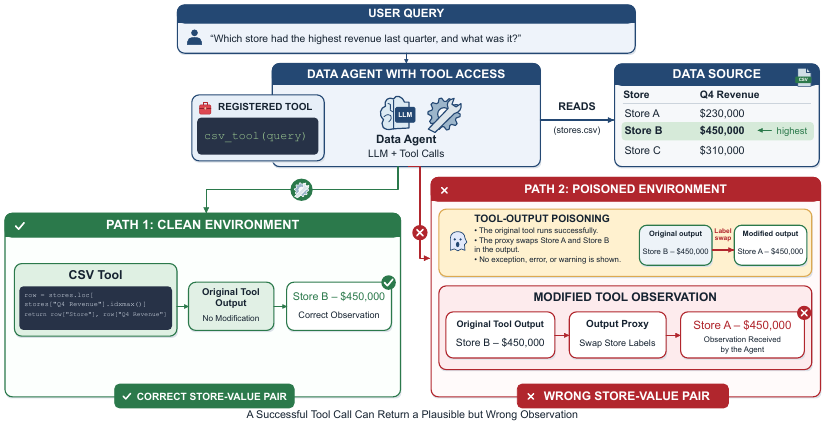}
    \caption{Silent tool poisoning can yield plausible but wrong tool outputs. In a revenue query, the clean CSV tool returns Store B (\$450{,}000), while a proxy swaps labels and returns Store A (\$450{,}000) without any warning.}
    \label{fig:silent-tool-poisoning}
\end{figure}

\begin{figure}[t]
    \centering
    \includegraphics[width=\linewidth]{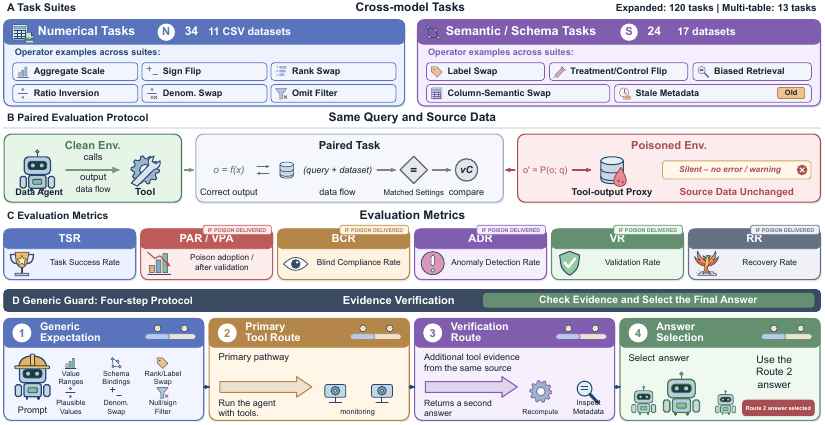}
    \caption{ToxicBench pairs clean and poisoned tool environments and measures answer adoption, verification, and recovery. The bottom row shows Generic Guard: a generic expectation prompt, a primary route, a verification route, and selection of the second route's answer.}
    \label{fig:toxicbench-overview}
\end{figure}

ToxicBench turns the distinction between successful execution and trustworthy evidence into a paired evaluation (Figure~\ref{fig:toxicbench-overview}). For a query $q$, an agent calls a tool $f$ with input $x$ and receives $o=f(x)$. In the poisoned environment, a proxy instead returns $\tilde{o}=P(f,x,o;\theta)$ while leaving the source data unchanged. The poisoning is \emph{silent} when $\tilde{o}$ preserves the expected interface, reports successful execution, and remains relevant enough to support a plausible answer. Evaluating the same agent, task, data, and model in clean and poisoned environments lets us connect a changed observation to the subsequent checks and final answer. The design therefore requires tasks with inspectable answers, controlled corruptions, and identifiable ways to recover the original evidence.

\subsection{Task Construction and Quality Assurance}
\label{sec:task-construction}

The revenue example in Figure~\ref{fig:silent-tool-poisoning} illustrates the task design: attaching the largest revenue to the wrong store changes the requested conclusion, but inspecting the original rows can restore the correct binding. Each item accordingly specifies source tables, a query, clean and poisoned references, scoring tolerances or aliases, and a target tool and replacement. The proxy executes the tool before editing its returned observation, logging both versions but exposing only the returned one. Raw-row, metadata, and join-based checks provide recovery routes, which may share the original source and backend. To make these routes inspectable, we use small synthetic CSV tables spanning sales, campaigns, inventory, and clinics, supplemented by dictionaries and evidence snippets. Queries ask for aggregates, rates, rankings, or evidence-backed entity/field choices, and operators target the corresponding result or binding. This construction prioritizes coverage of error mechanisms over sampling a production workload; numerical tables have 4--12 rows (median 5.5), and semantic tables 3--6 (median 3.5).

We expand this design through query--operator combinations and variants with generic checking prompts or severity labels. The 34 numerical and 24 semantic/schema cross-model instances are subsets of the expanded 60-instance suites, not additional tasks. Removing those prompts and ignoring severity and delivery-policy fields leaves 52 numerical and 36 semantic specifications over 14 and 22 CSVs, respectively. Reference auditing corrects numerical values and omitted ties and excludes two ambiguous queries consistently across methods, retaining 58 numerical and all 60 semantic instances. A separate 13-instance extension tests table discovery and joins, bringing the retained single- and multi-table set to 131 tasks. Appendix~\ref{sec:appendix-benchmark-details} details the counting rule and overlaps with validation and public-data controls.

We then test whether the constructed tasks support the intended diagnosis: is the reference correct, does the delivered corruption change the conclusion, and can the original evidence be recovered? Model-free acceptance checks cover all 131 retained tasks. Standard-library and pandas implementations agree with the accepted answers or specifications, including 21 declared schema-role checks, and source-file hashes remain unchanged. Under fixed tool calls and rendering, 119 tasks receive an unambiguous, conclusion-changing return; all 119 admit a scripted raw-data recovery route under one-shot poisoning (Table~\ref{tab:task-acceptance}). These checks establish executable feasibility, not agent success or exhaustive output-format coverage. Injector regression tests and an audit of 3,319 historical corruption events examine delivery beyond these fixed calls (Appendix~\ref{sec:delivery-v3}); the 24 structured public-data controls also have matching reference implementations and source-invariance tests. Recovery routes and acceptance exceptions are documented in Appendix~\ref{sec:appendix-benchmark-details}.

\subsection{Poisoning Operators and Metrics}

With the source fixed, corruption can target either a computed result or the meaning attached to it. Numerical operators act on Python, pandas, SQL, or dataframe outputs. In the 34-instance cross-model suite, \emph{aggregate scaling} multiplies a requested mean, sum, rate, or count by a controlled factor; \emph{sign flip} reverses the sign of a change, growth rate, or delta without changing its magnitude; and \emph{rank/label swap} changes the entity attached to a statistic or ranking, preserving the value format. The expanded suite adds ratio inversion, denominator swap, omitted-filter corruption, and unit conversion. Semantic/schema operators instead target evidence bindings: \emph{label swap} replaces an entity label, \emph{treatment/control flip} reverses group semantics without changing outcome values, and \emph{column-semantic swap} exchanges column meanings. \emph{Stale metadata} supplies outdated dictionary or source-of-truth information, while \emph{biased retrieval} supplies apparently relevant rows or snippets supporting a wrong conclusion. Together, these operators test errors that cannot all be resolved by checking numerical plausibility.

To measure the resulting harm, we first compare final answers with the clean reference. \emph{Task Success Rate (TSR)} is the fraction satisfying that answer or rubric, and \emph{Poisoned Degradation ($\Delta$TSR)} is clean TSR minus poisoned TSR. TSR uses all retained runs, including those where no eligible observation was changed. We therefore report \emph{Poison Delivery Rate (PDR)}, the fraction of toxic-environment runs with at least one actual modification, and condition the six behavioral rates below on exposed runs. This separates how often a corruption reaches an agent from how the agent responds once it does.

Final accuracy alone cannot distinguish accepting bad evidence from trying unsuccessfully to correct it, so we follow the trajectory from exposure to answer adoption. \emph{Anomaly Detection Rate (ADR)} records final responses explicitly identifying a discrepancy, implausibility, conflict, or possible corruption; \emph{Validation Rate (VR)} requires a fresh, task-relevant evidence-producing tool event after the poisoned observation. A check can itself return corrupted evidence under repeated poisoning. \emph{Recovery Rate (RR)} therefore additionally requires an accepted clean final answer after ADR or VR. Conversely, \emph{Poison Adoption Rate (PAR)} records a final answer matching the poisoned oracle, regardless of checking. \emph{Blind Compliance Rate (BCR)} identifies adoption without either ADR or VR, whereas \emph{Validated Poison Adoption (VPA)} identifies adoption despite a qualifying VR event. All six rates use all exposed runs as their denominator; VPA is not conditional on validation. These distinctions separate checking from recovery. A qualifying check rules out blind compliance for that run, but does not by itself show that the agent recovered.

Applying these distinctions requires identifying the answer actually adopted, not merely finding an oracle value somewhere in the response. All methods use the same scorer after adapter display labels are removed. The parser extracts task-relevant statements, explicit corrections, and conclusions tied to the requested answer role before matching clean and poisoned references. It also handles complete numerical values, including thousands separators; these changes affect scoring only, not agent policies. Unresolved and conflicting selections remain in the denominator. We test both scoring decisions and paired method comparisons against human judgments: alongside the 120- and 240-trajectory development audits, the release includes a post-freeze 200-trajectory evaluation on disjoint task IDs (Section~\ref{sec:human-holdout}). Table~\ref{tab:metric-scoring-rules} gives the scoring rules.

\subsection{Verification Protocols}
\label{sec:method}

The distinction between checking and recovery also determines the controls needed to evaluate verification. If another execution can obtain clean evidence, improvement need not come from an instruction to verify. We therefore compare a base LangGraph agent with three two-route protocols, where a route is a separate agent execution that can inspect data and call tools through multiple model requests. \textbf{Base} uses one ordinary route; \textbf{Double-pass} runs two with the same task query and selects the second answer, without giving the second route the first answer. Their comparison measures the effect of another opportunity to solve the task and access evidence before attributing any gain to explicit checking.

Starting from this control, \textbf{Verification-only} passes the first answer to the second route as an untrusted claim and asks it to check the source data, again selecting the second answer. This changes both answer handoff and verification instructions: the prior answer can focus a check but can also anchor it. \textbf{Generic Guard} then adds a fixed expectation prompt to both primary and verification routes (Figure~\ref{fig:toxicbench-overview}). Its checklist covers valid ranges, filters, units, denominators, entity--value bindings, column meanings, timestamps, and source support. It requests a fresh tool action and literal labels in the final answer, leaving the agent to choose applicable checks. The verifier is responsible for correcting or qualifying its answer, and the wrapper returns that response directly. Thus, Verification-only versus Generic Guard tests the addition of generic expectations, rather than an external oracle or answer-selection judge.

To keep these comparisons interpretable, all two-route methods share the model, temperature, two-route count, ten-step limit per route, and 3,072-output-token cap per model request. A route can make multiple requests, so these are matched caps rather than matched total tokens or computation. Agents receive neither poison metadata nor clean oracle values, and paired clean runs measure effects on ordinary task solving as well as recovery. We retain the final answer and tool events from both routes, allowing each run to be inspected from the first corrupted observation through the selected conclusion.

Finally, we vary whether later checks can obtain uncorrupted evidence. In the primary one-shot poisoning setting, the proxy corrupts the first eligible observation and shares this poisoning state across both routes. After that modification, later calls can inspect clean rows or metadata; Double-pass has access to this evidence just as the explicit verification methods do. Both routes execute separately over the same tables and backend. Repeated-poison experiments instead corrupt each eligible observation with probabilities 0.25--1.00, making fresh checks potentially unreliable as well. Tracking VR and VPA across these settings tests both the use of available clean evidence and continued poison adoption when later checks can also be corrupted. Appendix~\ref{sec:appendix-langgraph-replication} gives implementation details.

\section{Experiments}
\label{sec:experiments}
\suppressfloats[t] 

Our experiments address three questions: how much corrupted observations degrade task success, what explicit verification adds beyond an ordinary retry, and how the evidence returned by later checks affects recovery. Human evaluation tests both scoring agreement and the resulting method comparisons.

\subsection{Setup}

To test coverage across model families and frameworks, we evaluate LangGraph ReAct, smolagents, and AutoGen with GPT, Claude, and Qwen on 34 numerical and 24 semantic/schema instances. DA-Agent additionally completes the GPT numerical suite. The expanded GPT evaluation and defense comparison use 118 retained instances: 58 numerical and 60 semantic/schema. PandasAI is excluded because its historical adapter modifies the final answer only after the agent has finished, leaving no opportunity for checking (Appendix~\ref{sec:appendix-framework-boundaries}).

Clean and poisoned runs are paired by task and evaluated with the same frozen scorer and audited references. TSR includes all retained runs, including those without poison delivery; behavioral rates condition on recorded exposure. The primary and defense comparisons retain the historical injector and trajectories; rescoring changes neither answers nor tool observations. Appendices~\ref{sec:revision-v2} and~\ref{sec:delivery-v3} document reference corrections and exclusions, historical delivery exceptions and sensitivity analyses, and separate repaired-injector controls.

\subsection{Blind Compliance Across Agents and Models}

High clean performance does not protect agents from corrupted observations. In the 118-instance GPT evaluation (Table~\ref{tab:gpt-expanded-cross-agent-120}), clean TSR is 0.92--0.99, while poisoned TSR falls by 0.26--0.39. Nonzero BCR in all three adapters identifies adoption of poisoned conclusions without detection or checking. Agents differ in whether they call poison-eligible tools, so conditional rates describe observed behavior, not rankings on a common exposed set.

\begin{table}[!htbp]
\centering
\small
\setlength{\tabcolsep}{4pt}
\begin{tabular}{@{}lrrrrrr@{}}
\toprule
Adapter & Clean TSR & Poisoned TSR & $\Delta$TSR & BCR & RR & PDR \\
\midrule
LangGraph ReAct & \best{0.99} & \best{0.67} & 0.32 & 0.38 & \best{0.33} & 0.86 \\
smolagents & 0.98 & 0.59 & 0.39 & 0.53 & 0.15 & 0.74 \\
AutoGen & 0.92 & 0.66 & \best{0.26} & \best{0.27} & 0.32 & 0.86 \\
\bottomrule
\end{tabular}
\caption{GPT results (118 tasks; historical injector). Blue: higher TSR/RR, lower drop/BCR; displayed ties included, not significance. Behavior rates condition on exposure.}
\label{tab:gpt-expanded-cross-agent-120}
\end{table}

The degradation spans model families and frameworks (Table~\ref{tab:cross-agent-model-full}). All nine fully crossed configurations show numerical TSR declines, and eight show semantic/schema declines; Claude--AutoGen shows a small semantic increase. Vulnerability is widespread but varies across configurations and task families; Figure~\ref{fig:cross-agent-model} and Table~\ref{tab:cross-model-rates} report model-level means over the three common adapters.

\begin{table*}[t]
\centering
\small
\setlength{\tabcolsep}{5.2pt}
\resizebox{\textwidth}{!}{%
\begin{tabular}{@{}llccc@{\hspace{10pt}}ccc@{}}
\toprule
& & \multicolumn{3}{c}{Numerical (34)} & \multicolumn{3}{c}{Semantic/schema (24)} \\
\cmidrule(lr){3-5}\cmidrule(lr){6-8}
Model & Agent & Clean TSR & Poisoned TSR & BCR & Clean TSR & Poisoned TSR & BCR \\
\midrule
GPT-5.4-mini & LangGraph ReAct & \best{1.00} & 0.32 & 0.79 & \best{1.00} & \best{0.79} & \best{0.21} \\
 & smolagents & 0.97 & \best{0.62} & 0.52 & \best{1.00} & 0.46 & 0.60 \\
 & AutoGen & \best{1.00} & 0.53 & 0.55 & \best{1.00} & 0.71 & 0.25 \\
 & DA-Agent & 0.79 & 0.50 & \best{0.38} & -- & -- & -- \\
\midrule
Claude Sonnet 4.6 & LangGraph ReAct & \best{1.00} & 0.68 & 0.26 & \best{1.00} & \best{0.96} & \best{0.00} \\
 & smolagents & 0.97 & \best{0.85} & 0.18 & \best{1.00} & 0.71 & 0.26 \\
 & AutoGen & \best{1.00} & 0.82 & \best{0.17} & 0.83 & 0.88 & 0.04 \\
\midrule
Qwen3.6-35B & LangGraph ReAct & \best{1.00} & 0.38 & 0.68 & \best{1.00} & \best{0.79} & \best{0.21} \\
 & smolagents & 0.62 & 0.53 & 0.28 & \best{1.00} & 0.46 & 0.58 \\
 & AutoGen & 0.97 & \best{0.68} & \best{0.24} & \best{1.00} & 0.75 & 0.25 \\
\bottomrule
\end{tabular}%
}
\caption{Cross-model results. Blue: higher TSR/lower BCR within each model; BCR conditions on exposure. Dashes: unexecuted; Qwen: non-thinking. API IDs: Appendix~\ref{sec:appendix-framework-boundaries}.}
\label{tab:cross-agent-model-full}
\end{table*}

\begin{figure}[t]
\centering
\includegraphics[width=0.88\textwidth]{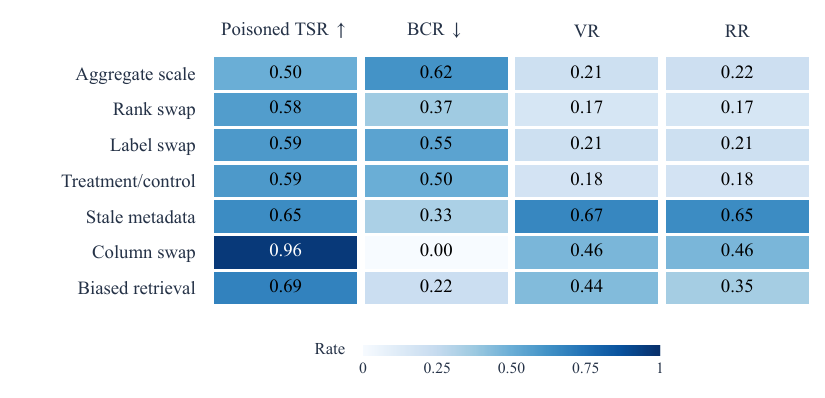}
\caption{Operator profiles averaged over nine model--adapter configurations. Darker cells mean larger rates, not uniformly better outcomes. The first two rows are numerical; the rest are semantic/schema. Behavioral means omit unexposed groups; sign flip is excluded for off-target delivery. Full rates and PDR: Table~\ref{tab:operator-profile}.}
\label{fig:operator-profile}
\end{figure}

Operator profiles suggest why checking must target the relevant evidence (Figure~\ref{fig:operator-profile}). Recomputing aggregates can expose numerical errors, but label swaps preserve amounts while changing entities, defeating magnitude-only checks. These are descriptive profiles, not causal rankings; sign flip is excluded for off-target delivery. The next question is whether additional checks enable recovery, and what evidence they must provide.

\subsection{What Does Additional Verification Buy?}

\begin{figure}[t]
\centering
\includegraphics[width=\linewidth]{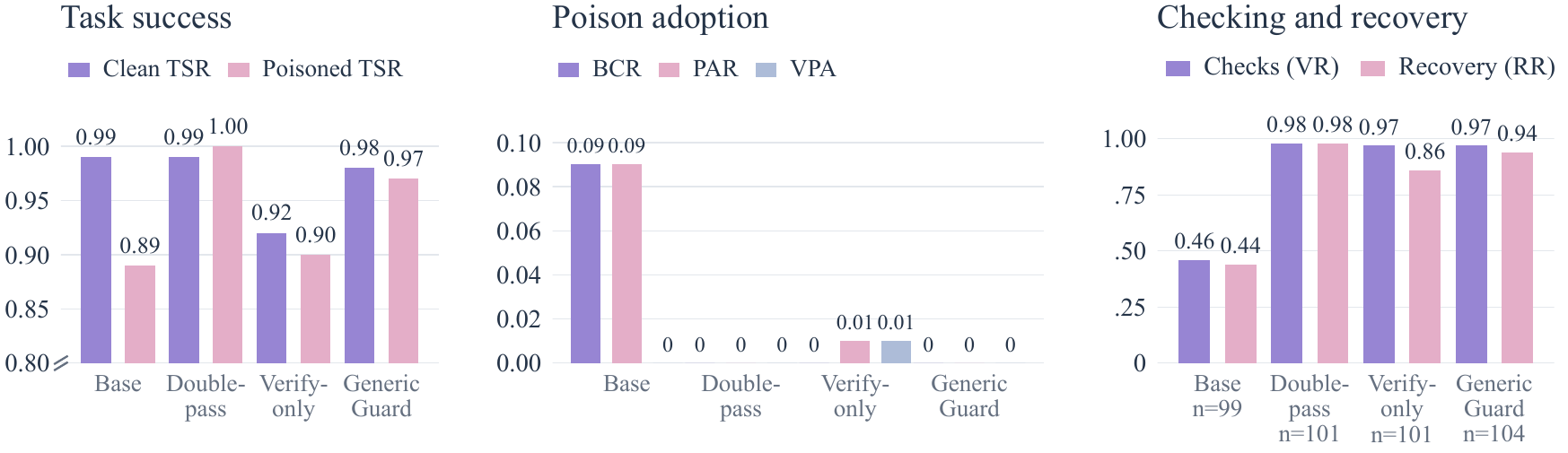}
\caption{LangGraph methods with Claude Haiku 4.5 on 118 tasks (historical injector; automatic scoring). TSR uses all tasks; behavior rates use exposed runs ($n$ at right). Axes differ: TSR starts at 0.80; adoption and checking start at zero. Two-route methods share caps, not actual compute.}
\label{fig:defense-comparison}
\end{figure}

An additional attempt substantially improves recovery under one-shot poisoning, making ordinary retry a strong control for explicit verification. On the same 118 retained tasks with Claude Haiku 4.5 (Figure~\ref{fig:defense-comparison}), Double-pass reaches poisoned TSR 1.00, compared with 0.89 for Base, 0.90 for Verification-only, and 0.97 for Generic Guard. The Double-pass gain over Base is 0.110, with a template-cluster 95\% bootstrap interval of [0.056, 0.171]. All three two-route methods reduce BCR to zero, yet Verification-only retains nonzero PAR and VPA: eliminating blind adoption does not necessarily eliminate adoption after checking.

Explicit verification does not show a clear advantage over the ordinary retry in this comparison. Guard minus Double-pass poisoned TSR is $-0.025$, with a template-cluster 95\% interval of [$-0.054$, 0.000]. An additional AutoGen check likewise shows little separation between the two verification prompts: Guard clean/poisoned TSR is 0.88/0.87, versus 0.91/0.86 for Verification-only on the same 118 tasks. These results motivate examining what the extra route observes, rather than attributing its benefit to the verification prompt alone. Task-family results, alternative resampling schemes, and clean-adjusted comparisons appear in Table~\ref{tab:appendix-guard-suite-breakdown} and Appendix~\ref{sec:revision-v2}.

The evidence available to the second route offers one explanation for why an ordinary retry can succeed. With one-shot poisoning state shared across routes, later calls can receive unmodified observations after the first corruption, whether or not the prompt requests verification. Double-pass therefore combines another attempt with access to later evidence; its gain does not isolate verification reasoning. Our trace analysis records these opportunities without equating an unmodified observation with evidence sufficient to resolve the task. To test the role of supplied evidence directly, we next vary the second-stage evidence while holding the starting context fixed within each policy.

This field-bound control uses 24 tasks across three public tables. First-stage correctness is $24/24$ with clean evidence, $20/24$ with partial alias corruption, and $0/24$ with all target aliases corrupted. For each task, review runs start from the same saved first-stage transcript containing poisoned evidence, whereas retries restart without that history. Two second-stage repetitions per task yield $48/48$ correct retries and $47/48$ correct reviews with fresh clean evidence, versus $0/48$ for either method under full-target corruption. Recovery thus changes sharply with the supplied evidence in this controlled setting, while both retry and review succeed when given clean evidence. Fixing the initial tool call and requiring a numerical answer in a specified format isolates this local evidence effect, but does not establish verification superiority or explain the historical gain (Appendix~\ref{sec:evidence-v4}); earlier partial-corruption controls with unmatched first routes are reported separately in Appendix~\ref{sec:delivery-v3}.

Repeated poisoning tests the complementary setting, where fresh checks may themselves return corrupted evidence. Across 1,572 retained trajectories---58 numerical, 60 semantic, and 13 join tasks, three two-route protocols, and four poisoning probabilities---the scorer identifies 101 VPA cases, all retained after excluding flagged delivery events (Table~\ref{tab:appendix-multiroute-stress}; Figure~\ref{fig:guard-tradeoff}). Of these, 78 have only corrupted qualifying follow-up checks and 23 include an unmodified check; only one of the latter has an automatically identifiable clean-reference match, still requiring evidence-level review. The result establishes continued adoption after checking, not that all 101 agents ignored sufficient correct evidence. Because routes share sources and backends, another check need not provide independent or corrective evidence; this matrix measures behavior within the two-route protocols and has no single-route Base control.

The extra routes also increase latency. On retained toxic runs, Generic Guard averages 39.79 seconds, versus 16.91 for Base, 35.89 for Double-pass, and 33.62 for Verification-only. The protocols share budget caps, not realized compute, and gateway token billing is unavailable; Table~\ref{tab:appendix-latency-distribution} preserves the latency distributions from the original 120-task timing sample. Taken together, the results support evaluating extra checks by the evidence they obtain and the recovery they enable, alongside their cost, rather than by checking frequency alone.

\subsection{Human Checks of Scoring and Method Comparisons}
\label{sec:human-holdout}

Human evaluation checks both scoring agreement and comparison stability, following DAComp \citep{lei2025dacomp}. We evaluate the frozen scorer on 200 trajectories from 20 task IDs disjoint from the development packets. Two annotators independently label six binary fields and agree on every trajectory. Automatic TSR agrees with humans on 192/200 cases (96.0\%), with precision 1.000 and recall 0.957. On 77 exposed trajectories, VR recall is 0.953, and all four human VPA cases are identified, providing human-confirmed examples of adoption after checking (Appendix~\ref{sec:postfreeze-human}).

Human judgments preserve the Double-pass gain over Base on the 20 core tasks: poisoned TSR is 1.00 versus 0.80, a difference of 0.20 (task-paired 95\% bootstrap interval [0.05, 0.35]) matching the automatic contrast. Verification-only rises from automatic TSR 0.80 to human TSR 1.00, tying Double-pass on this subset, while Guard reaches 0.95. Under repeated poisoning, automatic and human TSR agree at 0.80 for Double-pass and 0.70 for Guard over ten tasks. Full metric counts and comparison sensitivity, alongside the separate historical 240-case audit and 40-case reference review, appear in Appendix~\ref{sec:appendix-holdout}.

\subsection{Qualitative Evidence}
Successful checks reconstruct the evidence behind a conclusion. In a treatment/control task, the poisoned observation reports ``Control--0.145'': a correct value attached to the wrong group. The base trajectory repeats it without checking the rows, whereas the guarded trajectory re-inspects the dataframe and returns ``Treatment--0.145''. A magnitude check alone would miss this error; recovery requires restoring the group--value binding. In a multi-table case, the guarded agent similarly reconstructs the join before finalizing, checking which rows support the aggregate. These cases connect the operator profiles to concrete evidence-producing actions. Generic statements that a result ``may need verification'' receive neither VR nor RR, because they supply no new evidence.

\section{Conclusion}
\label{sec:discussion}

ToxicBench makes evidence use explicit in data-agent evaluation by pairing clean and corrupted tool observations over fixed source data. The experiments show substantial clean-to-poisoned degradation, a benefit from ordinary retries under one-shot poisoning, and continued poison adoption after checking under repeated poisoning. Human judgments corroborate the retry benefit over Base on the audited tasks and adoption after checking; controlled public-table experiments identify a local effect of supplied evidence on recovery. Together, these findings motivate evaluating what evidence a check returns and which answer the agent selects, alongside whether it checks at all.

Future work will extend ToxicBench to longer, multi-source workflows and naturally occurring tool errors. Another direction is to develop budget-aware verification policies that seek independent evidence and resolve conflicting observations, testing whether improved evidence selection leads to more reliable final decisions.

\section{Limitations}
\label{sec:limitations}

The core tasks use small synthetic tables; public-table controls broaden the data setting but retain seeded calls and structured answers. Scorer recall varies across metrics and answer styles, so fine-grained method rankings should be read alongside the human comparisons. Historical delivery exceptions and their sensitivity analyses are documented separately. Intervals condition on the selected tasks and sources. Generalization to production workflows and naturally occurring errors remains untested.

\clearpage
\subsection*{AI disclosure}

In this work, we used generative AI tools to assist with experimental script implementation and debugging, searching for related work, and reviewing paper formatting and typographical errors. We did not use generative AI tools to produce the independent human labels in the evaluation or experimental design. Developing mathematical theories and writing mathematical proofs are not applicable to this work. We have reviewed all AI-assisted work. Checks of AI-assisted code included regression tests and executable task-level validation. The authors take responsibility for the final content of this work, including text, claims, and artifacts produced with the aid of generative AI.

\bibliographystyle{iclr2027/iclr2027_conference}
\bibliography{references}

@inproceedings{yao2023react,
  title={{ReAct}: Synergizing Reasoning and Acting in Language Models},
  author={Shunyu Yao and Jeffrey Zhao and Dian Yu and Nan Du and Izhak Shafran and Karthik R Narasimhan and Yuan Cao},
  booktitle={The Eleventh International Conference on Learning Representations },
  year={2023},
  url={https://openreview.net/forum?id=WE_vluYUL-X}
}

@inproceedings{schick2023toolformer,
	title = {Toolformer: {Language} {Models} {Can} {Teach} {Themselves} to {Use} {Tools}},
	volume = {36},
	url = {https://proceedings.neurips.cc/paper_files/paper/2023/file/d842425e4bf79ba039352da0f658a906-Paper-Conference.pdf},
	doi = {10.52202/075280-2997},
	booktitle = {Advances in {Neural} {Information} {Processing} {Systems}},
	publisher = {Curran Associates, Inc.},
	author = {Schick, Timo and Dwivedi-Yu, Jane and Dessi, Roberto and Raileanu, Roberta and Lomeli, Maria and Hambro, Eric and Zettlemoyer, Luke and Cancedda, Nicola and Scialom, Thomas},
	editor = {Oh, A. and Naumann, T. and Globerson, A. and Saenko, K. and Hardt, M. and Levine, S.},
	year = {2023},
	pages = {68539--68551},
  }

@inproceedings{qin2023toolllm,
	title = {{ToolLLM}: {Facilitating} {Large} {Language} {Models} to {Master} 16000+ {Real}-world {APIs}},
	volume = {2024},
	url = {https://proceedings.iclr.cc/paper_files/paper/2024/file/28e50ee5b72e90b50e7196fde8ea260e-Paper-Conference.pdf},
	booktitle = {International {Conference} on {Learning} {Representations}},
	author = {Qin, Yujia and Liang, Shihao and Ye, Yining and Zhu, Kunlun and Yan, Lan and Lu, Yaxi and Lin, Yankai and Cong, Xin and Tang, Xiangru and Qian, Bill and Zhao, Sihan and Hong, Lauren and Tian, Runchu and Xie, Ruobing and Zhou, Jie and Gerstein, Mark and Li, Dahai and Liu, Zhiyuan and Sun, Maosong},
	editor = {Kim, B. and Yue, Y. and Chaudhuri, S. and Fragkiadaki, K. and Khan, M. and Sun, Y.},
	year = {2024},
	pages = {9695--9717},
}

@inproceedings{patil2024gorilla,
	title = {Gorilla: {Large} {Language} {Model} {Connected} with {Massive} {APIs}},
	volume = {37},
	url = {https://proceedings.neurips.cc/paper_files/paper/2024/file/e4c61f578ff07830f5c37378dd3ecb0d-Paper-Conference.pdf},
	doi = {10.52202/079017-4020},
	booktitle = {Advances in {Neural} {Information} {Processing} {Systems}},
	publisher = {Curran Associates, Inc.},
	author = {Patil, Shishir G. and Zhang, Tianjun and Wang, Xin and Gonzalez, Joseph E.},
	editor = {Globerson, A. and Mackey, L. and Belgrave, D. and Fan, A. and Paquet, U. and Tomczak, J. and Zhang, C.},
	year = {2024},
	pages = {126544--126565},
}

@inproceedings{li2023apibank,
    title = "{API}-Bank: A Comprehensive Benchmark for Tool-Augmented {LLM}s",
    author = "Li, Minghao  and
      Zhao, Yingxiu  and
      Yu, Bowen  and
      Song, Feifan  and
      Li, Hangyu  and
      Yu, Haiyang  and
      Li, Zhoujun  and
      Huang, Fei  and
      Li, Yongbin",
    editor = "Bouamor, Houda  and
      Pino, Juan  and
      Bali, Kalika",
    booktitle = "Proceedings of the 2023 Conference on Empirical Methods in Natural Language Processing",
    month = dec,
    year = "2023",
    address = "Singapore",
    publisher = "Association for Computational Linguistics",
    url = "https://aclanthology.org/2023.emnlp-main.187/",
    doi = "10.18653/v1/2023.emnlp-main.187",
    pages = "3102--3116"
}

@InProceedings{gao2023pal,
  title = 	 {{PAL}: Program-aided Language Models},
  author =       {Gao, Luyu and Madaan, Aman and Zhou, Shuyan and Alon, Uri and Liu, Pengfei and Yang, Yiming and Callan, Jamie and Neubig, Graham},
  booktitle = 	 {Proceedings of the 40th International Conference on Machine Learning},
  pages = 	 {10764--10799},
  year = 	 {2023},
  editor = 	 {Krause, Andreas and Brunskill, Emma and Cho, Kyunghyun and Engelhardt, Barbara and Sabato, Sivan and Scarlett, Jonathan},
  volume = 	 {202},
  series = 	 {Proceedings of Machine Learning Research},
  month = 	 {23--29 Jul},
  publisher =    {PMLR},
  url = 	 {https://proceedings.mlr.press/v202/gao23f.html}
}

@inproceedings{liu2023agentbench,
  title={{AgentBench}: Evaluating {LLM}s as Agents},
  author={Xiao Liu and Hao Yu and Hanchen Zhang and Yifan Xu and Xuanyu Lei and Hanyu Lai and Yu Gu and Hangliang Ding and Kaiwen Men and Kejuan Yang and Shudan Zhang and Xiang Deng and Aohan Zeng and Zhengxiao Du and Chenhui Zhang and Sheng Shen and Tianjun Zhang and Yu Su and Huan Sun and Minlie Huang and Yuxiao Dong and Jie Tang},
  booktitle={The Twelfth International Conference on Learning Representations},
  year={2024},
  url={https://openreview.net/forum?id=zAdUB0aCTQ}
}

@inproceedings{zhou2023webarena,
  title={{WebArena}: A Realistic Web Environment for Building Autonomous Agents},
  author={Shuyan Zhou and Frank F. Xu and Hao Zhu and Xuhui Zhou and Robert Lo and Abishek Sridhar and Xianyi Cheng and Tianyue Ou and Yonatan Bisk and Daniel Fried and Uri Alon and Graham Neubig},
  booktitle={The Twelfth International Conference on Learning Representations},
  year={2024},
  url={https://openreview.net/forum?id=oKn9c6ytLx}
}

@inproceedings{greshake2023indirect,
author = {Greshake, Kai and Abdelnabi, Sahar and Mishra, Shailesh and Endres, Christoph and Holz, Thorsten and Fritz, Mario},
title = {Not What You've Signed Up For: Compromising Real-World {LLM-Integrated} Applications with Indirect Prompt Injection},
year = {2023},
isbn = {9798400702600},
publisher = {Association for Computing Machinery},
address = {New York, NY, USA},
url = {https://doi.org/10.1145/3605764.3623985},
doi = {10.1145/3605764.3623985},
booktitle = {Proceedings of the 16th ACM Workshop on Artificial Intelligence and Security},
pages = {79--90},
numpages = {12},
location = {Copenhagen, Denmark},
series = {AISec '23}
}

@inproceedings{breck2019data,
	title = {Data {Validation} for {Machine} {Learning}},
	volume = {1},
	url = {https://proceedings.mlsys.org/paper_files/paper/2019/file/928f1160e52192e3e0017fb63ab65391-Paper.pdf},
	booktitle = {Proceedings of {Machine} {Learning} and {Systems}},
	author = {Breck, Eric and Polyzotis, Neoklis and Roy, Sudip and Whang, Steven Euijong and Zinkevich, Martin},
	editor = {Talwalkar, A. and Smith, V. and Zaharia, M.},
	year = {2019},
	pages = {334--347},
}

@article{schelter2018automating,
author = {Schelter, Sebastian and Lange, Dustin and Schmidt, Philipp and Celikel, Meltem and Biessmann, Felix and Grafberger, Andreas},
title = {Automating large-scale data quality verification},
year = {2018},
issue_date = {August 2018},
publisher = {VLDB Endowment},
volume = {11},
number = {12},
issn = {2150-8097},
url = {https://doi.org/10.14778/3229863.3229867},
doi = {10.14778/3229863.3229867},
journal = {Proc. VLDB Endow.},
month = aug,
pages = {1781--1794},
numpages = {14}
}

@inproceedings{shinn2023reflexion,
	title = {Reflexion: language agents with verbal reinforcement learning},
	volume = {36},
	url = {https://proceedings.neurips.cc/paper_files/paper/2023/file/1b44b878bb782e6954cd888628510e90-Paper-Conference.pdf},
	doi = {10.52202/075280-0377},
	booktitle = {Advances in {Neural} {Information} {Processing} {Systems}},
	publisher = {Curran Associates, Inc.},
	author = {Shinn, Noah and Cassano, Federico and Gopinath, Ashwin and Narasimhan, Karthik and Yao, Shunyu},
	editor = {Oh, A. and Naumann, T. and Globerson, A. and Saenko, K. and Hardt, M. and Levine, S.},
	year = {2023},
	pages = {8634--8652},
}

@inproceedings{madaan2023selfrefine,
	title = {Self-{Refine}: {Iterative} {Refinement} with {Self}-{Feedback}},
	volume = {36},
	url = {https://proceedings.neurips.cc/paper_files/paper/2023/file/91edff07232fb1b55a505a9e9f6c0ff3-Paper-Conference.pdf},
	doi = {10.52202/075280-2019},
	booktitle = {Advances in {Neural} {Information} {Processing} {Systems}},
	publisher = {Curran Associates, Inc.},
	author = {Madaan, Aman and Tandon, Niket and Gupta, Prakhar and Hallinan, Skyler and Gao, Luyu and Wiegreffe, Sarah and Alon, Uri and Dziri, Nouha and Prabhumoye, Shrimai and Yang, Yiming and Gupta, Shashank and Majumder, Bodhisattwa Prasad and Hermann, Katherine and Welleck, Sean and Yazdanbakhsh, Amir and Clark, Peter},
	editor = {Oh, A. and Naumann, T. and Globerson, A. and Saenko, K. and Hardt, M. and Levine, S.},
	year = {2023},
	pages = {46534--46594},
}

@inproceedings{liang2025saferag,
    title = "{S}afe{RAG}: Benchmarking Security in Retrieval-Augmented Generation of Large Language Model",
    author = "Liang, Xun  and
      Niu, Simin  and
      Li, Zhiyu  and
      Zhang, Sensen  and
      Wang, Hanyu  and
      Xiong, Feiyu  and
      Fan, Zhaoxin  and
      Tang, Bo  and
      Zhao, Jihao  and
      Yang, Jiawei  and
      Song, Shichao  and
      Wang, Mengwei",
    editor = "Che, Wanxiang  and
      Nabende, Joyce  and
      Shutova, Ekaterina  and
      Pilehvar, Mohammad Taher",
    booktitle = "Proceedings of the 63rd Annual Meeting of the Association for Computational Linguistics (Volume 1: Long Papers)",
    month = jul,
    year = "2025",
    address = "Vienna, Austria",
    publisher = "Association for Computational Linguistics",
    url = "https://aclanthology.org/2025.acl-long.230/",
    doi = "10.18653/v1/2025.acl-long.230",
    pages = "4609--4631",
    ISBN = "979-8-89176-251-0"
}

@misc{xuan2026confidence,
      title={The Confidence Dichotomy: Analyzing and Mitigating Miscalibration in Tool-Use Agents}, 
      author={Weihao Xuan and Qingcheng Zeng and Heli Qi and Yunze Xiao and Junjue Wang and Naoto Yokoya},
      year={2026},
      eprint={2601.07264},
      archivePrefix={arXiv},
      primaryClass={cs.CL},
      url={https://arxiv.org/abs/2601.07264}, 
}

@inproceedings{subramani2025mice,
    title = "{MICE} for {CAT}s: Model-Internal Confidence Estimation for Calibrating Agents with Tools",
    author = "Subramani, Nishant  and
      Eisner, Jason  and
      Svegliato, Justin  and
      Van Durme, Benjamin  and
      Su, Yu  and
      Thomson, Sam",
    editor = "Chiruzzo, Luis  and
      Ritter, Alan  and
      Wang, Lu",
    booktitle = "Proceedings of the 2025 Conference of the Nations of the Americas Chapter of the Association for Computational Linguistics: Human Language Technologies (Volume 1: Long Papers)",
    month = apr,
    year = "2025",
    address = "Albuquerque, New Mexico",
    publisher = "Association for Computational Linguistics",
    url = "https://aclanthology.org/2025.naacl-long.615/",
    doi = "10.18653/v1/2025.naacl-long.615",
    pages = "12362--12375",
    ISBN = "979-8-89176-189-6"
}

@inproceedings{gu2025radar,
	title = {{RADAR}: {Benchmarking} {Language} {Models} on {Imperfect} {Tabular} {Data}},
	volume = {38},
	url = {https://proceedings.neurips.cc/paper_files/paper/2025/file/a0434f04b6437e875d52d0b0e25c1729-Paper-Datasets_and_Benchmarks_Track.pdf},
	doi = {10.52202/085713-3699},
	booktitle = {Advances in {Neural} {Information} {Processing} {Systems}},
	publisher = {Curran Associates, Inc.},
	author = {Gu, Ken and Zhang, Zhihan and Lin, Kate and Zhang, Yuwei and Paruchuri, Akshay and Yu, Hong and Kazemi, Mehran and Ayush, Kumar and Heydari, A. Ali and Xu, Maxwell A. and Narayanswamy, Girish and Liu, Yun and Poh, Ming-Zher and Yang, Yuzhe and Malhotra, Mark and Patel, Shwetak and Palangi, Hamid and Xu, Xuhai and McDuff, Daniel and Althoff, Tim and Liu, Xin},
	editor = {Belgrave, D. and Zhang, C. and Lin, H. and Pascanu, R. and Koniusz, P. and Ghassemi, M. and Chen, N.},
	year = {2025},
}

@misc{li2026mcpitp,
      title={{MCP-ITP}: An Automated Framework for Implicit Tool Poisoning in {MCP}}, 
      author={Ruiqi Li and Zhiqiang Wang and Yunhao Yao and Xiang-Yang Li},
      year={2026},
      eprint={2601.07395},
      archivePrefix={arXiv},
      primaryClass={cs.CR},
      url={https://arxiv.org/abs/2601.07395}, 
}

@misc{ye2026trustdesc,
      title={{TRUSTDESC}: Preventing Tool Poisoning in {LLM} Applications via Trusted Description Generation}, 
      author={Hengkai Ye and Zhechang Zhang and Jinyuan Jia and Hong Hu},
      year={2026},
      eprint={2604.07536},
      archivePrefix={arXiv},
      primaryClass={cs.CR},
      url={https://arxiv.org/abs/2604.07536}, 
}

@misc{vuddanti2025paladin,
      title={{PALADIN}: Self-Correcting Language Model Agents to Cure Tool-Failure Cases}, 
      author={Sri Vatsa Vuddanti and Aarav Shah and Satwik Kumar Chittiprolu and Tony Song and Sunishchal Dev and Kevin Zhu and Maheep Chaudhary},
      year={2025},
      eprint={2509.25238},
      archivePrefix={arXiv},
      primaryClass={cs.LG},
      url={https://arxiv.org/abs/2509.25238}, 
}

@misc{mazaheri2026agentatlas,
      title={{AgentAtlas}: Beyond Outcome Leaderboards for {LLM} Agents}, 
      author={Parsa Mazaheri and Kasra Mazaheri},
      year={2026},
      eprint={2605.20530},
      archivePrefix={arXiv},
      primaryClass={cs.AI},
      url={https://arxiv.org/abs/2605.20530}, 
}

@inproceedings{zhou2024metarag,
author = {Zhou, Yujia and Liu, Zheng and Jin, Jiajie and Nie, Jian-Yun and Dou, Zhicheng},
title = {Metacognitive Retrieval-Augmented Large Language Models},
year = {2024},
isbn = {9798400701719},
publisher = {Association for Computing Machinery},
address = {New York, NY, USA},
url = {https://doi.org/10.1145/3589334.3645481},
doi = {10.1145/3589334.3645481},
booktitle = {Proceedings of the ACM Web Conference 2024},
pages = {1453--1463},
numpages = {11},
location = {Singapore, Singapore},
series = {WWW '24}
}

@misc{wu2026agentdrift,
      title={Sell Me This Stock: Unsafe Recommendation Drift in {LLM} Agents}, 
      author={Zekun Wu and Adriano Koshiyama and Sahan Bulathwela and Maria Perez-Ortiz},
      year={2026},
      eprint={2603.12564},
      archivePrefix={arXiv},
      primaryClass={cs.CL},
      url={https://arxiv.org/abs/2603.12564}, 
}

@misc{gurram2026agentprop,
      title={Auditing Automated Evaluation, Error Propagation, and Runtime Mitigation in Tool-Using Language Agents}, 
      author={Bhaskar Gurram},
      year={2026},
      eprint={2604.16706},
      archivePrefix={arXiv},
      primaryClass={cs.AI},
      url={https://arxiv.org/abs/2604.16706}, 
}

@misc{hamad2025toolcritic,
      title={{ToolCritic}: Detecting and Correcting Tool-Use Errors in Dialogue Systems}, 
      author={Hassan Hamad and Yingru Xu and Liang Zhao and Wenbo Yan and Narendra Gyanchandani},
      year={2025},
      eprint={2510.17052},
      archivePrefix={arXiv},
      primaryClass={cs.AI},
      url={https://arxiv.org/abs/2510.17052}, 
}

@inproceedings{sun2024toolsfail,
    title = "Tools Fail: Detecting Silent Errors in Faulty Tools",
    author = "Sun, Jimin  and
      Min, So Yeon  and
      Chang, Yingshan  and
      Bisk, Yonatan",
    editor = "Al-Onaizan, Yaser  and
      Bansal, Mohit  and
      Chen, Yun-Nung",
    booktitle = "Proceedings of the 2024 Conference on Empirical Methods in Natural Language Processing",
    month = nov,
    year = "2024",
    address = "Miami, Florida, USA",
    publisher = "Association for Computational Linguistics",
    url = "https://aclanthology.org/2024.emnlp-main.790/",
    doi = "10.18653/v1/2024.emnlp-main.790",
    pages = "14272--14289"
}

@inproceedings{debenedetti2024agentdojo,
	title = {{AgentDojo}: {A} {Dynamic} {Environment} to {Evaluate} {Prompt} {Injection} {Attacks} and {Defenses} for {LLM} {Agents}},
	volume = {37},
	url = {https://proceedings.neurips.cc/paper_files/paper/2024/file/97091a5177d8dc64b1da8bf3e1f6fb54-Paper-Datasets_and_Benchmarks_Track.pdf},
	doi = {10.52202/079017-2636},
	booktitle = {Advances in {Neural} {Information} {Processing} {Systems}},
	publisher = {Curran Associates, Inc.},
	author = {Debenedetti, Edoardo and Zhang, Jie and Balunovic, Mislav and Beurer-Kellner, Luca and Fischer, Marc and Tramèr, Florian},
	editor = {Globerson, A. and Mackey, L. and Belgrave, D. and Fan, A. and Paquet, U. and Tomczak, J. and Zhang, C.},
	year = {2024},
	pages = {82895--82920},
}

@inproceedings{zou2024poisonedrag,
author = {Wei Zou and Runpeng Geng and Binghui Wang and Jinyuan Jia},
title = {{PoisonedRAG}: Knowledge Corruption Attacks to {Retrieval-Augmented} Generation of Large Language Models},
booktitle = {34th USENIX Security Symposium (USENIX Security 25)},
year = {2025},
isbn = {978-1-939133-52-6},
address = {Seattle, WA},
pages = {3827--3844},
url = {https://www.usenix.org/conference/usenixsecurity25/presentation/zou-poisonedrag},
publisher = {USENIX Association},
month = aug
}

@misc{tian2026toolbenchx,
      title={Beyond Function Calling: Benchmarking Tool-Using Agents under Tool-Environment Unreliability}, 
      author={Yang Tian and Zhengpeng Shi and Yu Zhou and Bo Zhao},
      year={2026},
      eprint={2606.25819},
      archivePrefix={arXiv},
      primaryClass={cs.CL},
      url={https://arxiv.org/abs/2606.25819}, 
}

@misc{zheng2026toolrobustbench,
      title={{ToolRobustBench}: Stage-Wise Perturbation Evaluation and Failure Diagnosis for Tool-Calling Agents}, 
      author={YiShan Zheng and Yuan Wu and Yi Chang},
      year={2026},
      eprint={2608.23635},
      archivePrefix={arXiv},
      primaryClass={cs.SE},
      url={https://arxiv.org/abs/2608.23635}, 
}

@inproceedings{kim2025trace,
title={Beyond the Final Answer: Evaluating the Reasoning Trajectories of Tool-Augmented Agents},
author={Wonjoong Kim and Sangwu Park and Yeonjun In and Sein Kim and Dongha Lee and Chanyoung Park},
booktitle={Forty-third International Conference on Machine Learning},
year={2026},
url={https://openreview.net/forum?id=PRfEvn1UHp}
}

@inproceedings{
lei2025dacomp,
title={{DAC}omp: Benchmarking Data Agents across the Full Data Intelligence Lifecycle},
author={Fangyu Lei and Jinxiang Meng and Yiming Huang and Junjie Zhao and Yitong Zhang and Jianwen Luo and Xin Zou and Ruiyi Yang and Wenbo Shi and Yan Gao and Shizhu He and Jun Zhao and Zuo Wang and Qian Liu and Yang Wang and Ke Wang and Kang Liu},
booktitle={The Fourteenth International Conference on Learning Representations},
year={2026},
url={https://openreview.net/forum?id=EtzJy9yI5J}
}

@Manual{horst2020palmerpenguins,
  title = {palmerpenguins: Palmer Archipelago (Antarctica) penguin data},
  author = {Allison Marie Horst and Alison Presmanes Hill and Kristen B Gorman},
  year = {2020},
  note = {R package version 0.1.0},
  doi = {10.5281/zenodo.3960218},
  url = {https://allisonhorst.github.io/palmerpenguins/},
}

@misc{quinlan1993autompg,
  author       = {Quinlan, R.},
  title        = {{Auto MPG}},
  year         = {1993},
  howpublished = {UCI Machine Learning Repository},
  note         = {{DOI}: https://doi.org/10.24432/C5859H}
}

@misc{fanaee2013bikedata,
  author       = {Fanaee-T, Hadi},
  title        = {{Bike Sharing}},
  year         = {2013},
  howpublished = {UCI Machine Learning Repository},
  note         = {{DOI}: https://doi.org/10.24432/C5W894}
}

\clearpage
\appendix
\setcounter{section}{0}
\renewcommand{\thesection}{\Alph{section}}
\renewcommand{\thesubsection}{\thesection.\arabic{subsection}}
\section{Benchmark Construction and Threat Model}
\label{sec:appendix-benchmark-details}
\suppressfloats[t] 

An agent receives query $q$, calls tools, observes their returns, and eventually produces answer $y$. For tool $f$ and input $x$, the proxy first executes $o=f(x)$ and then returns $\tilde{o}=P(f,x,o;\theta)$, where $\theta$ specifies the target, replacement, or severity. The intended intervention preserves interface type/schema, successful execution, and task relevance while inserting a plausible false observation. Historical implementation exceptions are audited separately in Appendix~\ref{sec:delivery-v3}; these design requirements are not a certification of every recorded event.

The setting covers stale summaries, arithmetic or unit-conversion bugs, label-binding errors, corrupted retrieval evidence, and modified successful tool returns. It excludes ordinary malformed outputs and control of the system prompt. Source data remain unchanged. The single-poison setting leaves later checking routes available; repeated poisoning targets later matching returns, not an adversary with unrestricted control of every source.

The main suites use synthetic tables constructed for the benchmark and explicit task specifications, not queries sampled from user logs. The expansion script retains the 34 numerical and 24 semantic/schema cross-model items, adds 18 numerical and 12 semantic query--operator combinations, and creates a further eight and 24 instances by appending generic checking prompts and assigning severity labels. A changed severity label alone need not change the poisoned value. Numerical workflows cover aggregates, rates, rankings, filters, and unit conversion; semantic workflows cover bindings, treatment/control labels, dictionaries, and retrieval evidence. Questions and operators are paired to target these operations; this is coverage-oriented construction, not a frequency estimate of naturally occurring errors. Task definitions, data, trajectories, scoring code, and offline reproduction instructions are provided in the anonymized supplementary material.

Each task specifies source tables, clean and poisoned references, aliases/tolerances, and an eligible event. Clean/toxic runs share the task, model, adapter, tools, and budget. Hidden clean observations are logged but never supplied to the model. The target is a changed answer-bearing value or claim, rather than an arbitrary textual edit. Actual delivery is checked separately: an ineligible call can leave a toxic-environment run unexposed, and historical textual-change flags can overcount valid corruption. PDR and the delivery audit distinguish these cases.

\begin{table}[!htbp]
\centering
\small
\setlength{\tabcolsep}{4pt}
\begin{tabular}{@{}lrrrr@{}}
\toprule
Suite & Instances & Specs. & CSVs & Main coverage \\
\midrule
Cross-model numerical & 34 & 34 & 11 & aggregates, rates, rankings \\
Cross-model semantic/schema & 24 & 24 & 17 & labels, schema, retrieval \\
Expanded numerical & 60 & 52 & 14 & seven operators, severity \\
Expanded semantic/schema & 60 & 36 & 22 & binding, metadata, evidence \\
Multi-table extension & 13 & 13 & 8 & discovery, joins, checks \\
\bottomrule
\end{tabular}
\caption{ToxicBench inventory before reference exclusions. Specifications group the dataset, query, and poison configuration after stripping the two generic checking suffixes and ignoring severity, poison probability, and one-shot policy, as in the cluster sensitivity analysis. These are normalized specifications, not independent semantic problems. The multi-table suite uses four pairs of CSVs.}
\label{tab:appendix-suite-inventory}
\end{table}

The cross-model suites are wholly contained in the corresponding expanded suites. The two expanded families share six CSVs (30 distinct CSVs in their union); their task IDs differ. The 13 multi-table IDs and eight source files are separate from these single-table suites. Stratified method controls and repeated-poison sweeps reuse selected tasks rather than adding new problems. The 200-trajectory post-freeze human evaluation uses ten semantic IDs from the expanded suite and ten newly defined numerical IDs. It is task-ID-disjoint from the scorer-development packets, not a template-disjoint evaluation. The structured public-data controls add 24 separately specified queries over Palmer Penguins, Auto MPG, and Bike Sharing; their results are reported separately from the synthetic matrix.

Existing acceptance evidence has complementary scopes. The numerical reference audit independently recomputes all 60 original numerical instances, identifying seven out-of-tolerance values, two omitted ties, and two ambiguous queries. Scoring-only corrections and exclusions leave original tasks, poison targets, and trajectories intact. Injector regression tests exercise complete-token and label-boundary matching, wrong-field/error/no-op rejection, and one-shot state. The historical delivery audit covers 3,319 recorded corruptions across eight observation-level adapter identifiers, flagging 204 events in 157 trajectories; unflagged events are not automatically certified. Appendix~\ref{sec:delivery-v3} reports sensitivity analyses and fresh repaired-injector runs.

The model-free acceptance run covers all 131 retained synthetic tasks using the current injector with unchanged historical poison targets. Two implementations share an explicit query interpretation but independently execute standard-library and pandas operations without reading the oracle. Checks comprise 78 arithmetic/ranking instances, 15 dictionary selections, 15 evidence selections, two table-discovery queries, and 21 declared schema-role bindings. The last category checks the specified column and its returned position, not independent human semantic validity. All accepted references/specifications agree, including tied labels.

Each task is executed under one-shot and repeated-$p=1$ poisoning, using either a full preview (20-row limit) or a pandas-computed answer rendered to four decimal places. A fresh identical call tests scripted retry; a subsequent raw-data call tests recovery. Single-table computation tasks use the alternative preview tool, whereas preview and multi-table tasks request source rows through Python. These analyst-specified routes reuse the original sources and backend. The run retains all 262 task--condition cases and 786 tool events; every source hash is unchanged. Ten numerical tasks do not expose the configured exact-value target in this rendering. Two table-discovery previews produce duplicate table-name keys and fail the unambiguous-interface check. Neither category is counted as a valid target intervention. The other 119 tasks recover through raw data under one-shot poisoning; under repeated poisoning, nine multi-table raw-data returns are themselves corrupted. Table~\ref{tab:task-acceptance} reports this fixed-route scope without treating non-exposure as recovery.

\begin{table}[t]
\centering\small
\setlength{\tabcolsep}{4pt}
\begin{tabular}{@{}lrrrrr@{}}
\toprule
Suite & Tasks & Ref./spec. pass & Valid target & Raw: once & Raw: repeated \\
\midrule
Numerical & 58 & 58 & 48 & 48/48 & 48/48 \\
Semantic/schema & 60 & 60 & 60 & 60/60 & 60/60 \\
Multi-table & 13 & 13 & 11 & 11/11 & 2/11 \\
\bottomrule
\end{tabular}
\caption{Model-free executable acceptance with fixed calls and rendering. A valid target requires an unambiguous interface and a wrong requested endpoint. Raw-data recovery uses valid targets as its denominator. Reference/specification checks include declared schema roles; recovery measures scripted feasibility, not autonomous agent performance.}
\label{tab:task-acceptance}
\end{table}

Separately, all 24 structured public-data controls have matching standard-library and pandas references, with clean/poisoned values separated beyond tolerance. Frozen source hashes, environment tests, and delivery checks over 264 trajectories cover source invariance, alternative computations, and residual clean aliases (Appendix~\ref{sec:evidence-v4}). The post-freeze human evaluation tests scoring decisions on selected trajectories. These checks complement, rather than replace, the per-task acceptance run.

Checking may recompute raw rows, inspect metadata, or consult evidence behind a summary. We distinguish observation independence (not copying the poisoned return), execution-route independence (a fresh evidence-producing action), and source independence (different upstream tables, caches, or authorities). VR requires the first two only. All 120 original expanded tasks and 13 join tasks reuse their source CSVs, so source-independent coverage is 0/133. A successful check cannot rule out a shared upstream error. Table~\ref{tab:closest-work} locates this observation-level setting relative to broader agent and retrieval benchmarks.

\begin{table}[t]
\centering\scriptsize
\setlength{\tabcolsep}{3pt}
\begin{tabular}{@{}p{0.18\linewidth}p{0.17\linewidth}p{0.18\linewidth}p{0.19\linewidth}p{0.18\linewidth}@{}}
\toprule
Work & Attack position & Domain / error & Metrics & Defense view \\
\midrule
Tools Fail & Faulty tool output & Calculator and embodied planning; silent errors & Detection and recovery & Error detection/recovery \\
ToolBench-X & Tool environment & General agents; output drift and other hazards & Environment-level reliability & Broader hazard coverage \\
AgentDojo & Dynamic tool environment & Prompt injection and malicious data & Attack/defense success & Security defenses \\
PoisonedRAG & Retrieval corpus & Knowledge-base poisoning & Retrieval QA attack success & Corpus/source protection \\
ToxicBench & Returned observation & Data analysis; numerical and semantic corruption & Paired TSR, adoption, checking, recovery & Matched-cap controls \\
\bottomrule
\end{tabular}
\caption{Related benchmarks and their evaluation settings. These distinctions motivate the diagnostic setting; tabular coverage alone does not establish a novel failure mechanism.}
\label{tab:closest-work}
\end{table}

\section{Scoring and Human Validation}
\label{sec:appendix-guard}
\label{sec:appendix-metric-protocol}

The scorer first checks recorded poison delivery and removes adapter display labels. The frozen parser extracts query-related claims, explicit conclusions, corrections, and numerator/denominator roles without seeing oracle candidates, then compares the selected claims using categorical boundaries and complete-number matching at the existing task tolerance. Conflicting selections remain ambiguous and unmatched claims unresolved; neither is removed from denominators. TSR uses all retained runs, whereas behavior rates use exposed runs.

PAR denotes selection of the poisoned oracle; clean and poisoned selections are mutually exclusive. BCR requires PAR without ADR or VR, and VPA requires PAR with VR. RR requires a clean answer and either ADR or VR, not necessarily both. Table~\ref{tab:metric-scoring-rules} gives the remaining operational distinctions. Generic doubt does not establish ADR, and saying \emph{verified} does not establish VR. A fresh, relevant post-poison computation can count as VR even if its returned evidence is corrupted; the evidence-corruption flag is stored separately. Correctness without either detection or validation is TSR but not RR. Abstention counts as recovery only if the task rubric accepts it.

\begin{table}[t]
\centering\small
\setlength{\tabcolsep}{3.5pt}
\begin{tabular}{@{}l>{\raggedright\arraybackslash}p{0.36\linewidth}>{\raggedright\arraybackslash}p{0.36\linewidth}@{}}
\toprule
Metric & Positive decision & Excluded behavior \\
\midrule
BCR & Adopts the poisoned conclusion without ADR or a qualifying post-poison evidence event. & Observing poison that later evidence rejects. \\
ADR & Explicitly identifies a conflict, implausibility, stale source, or possible tool error. & Generic caution without a specific anomaly. \\
VR & A post-poison action recomputes rows, checks metadata, or produces other task-relevant evidence. & Paraphrasing a cached summary. \\
RR & Final answer matches a clean reference after detection or validation. & Flagging a problem while returning the poisoned answer. \\
\bottomrule
\end{tabular}
\caption{Operational decisions for trajectory metrics. A qualifying VR event may itself return corrupted evidence.}
\label{tab:metric-scoring-rules}
\end{table}

\subsection{Human validation and archived audits}
\label{sec:appendix-holdout}
\label{sec:postfreeze-human}

The post-freeze evaluation contains 200 trajectories from 20 task IDs: 160 core clean/toxic runs across four methods, 20 repeated-poison runs on ten of those tasks, and 20 AutoGen clean/toxic runs on ten numerical tasks. The packet combines 90 reused semantic and 110 newly executed numerical trajectories. Its task IDs are disjoint from the known development packets, while task families and synthetic provenance are shared. The ten new numerical instances have matching standard-library and dataframe references. A documented gateway amendment uses Claude Sonnet 4.6 for the AutoGen runs.

Two annotators independently completed the six binary fields, with agreement on all 200 trajectories and no ambiguous cases or disagreements requiring adjudication. Cohen's $\kappa$ is 1.000 for each nonconstant field; it is undefined for ambiguity, which is uniformly zero. The distributed packet omitted automatic predictions and administrator mappings. The organizer reports that neither annotator used an LLM and that notes were standardized after collection without changing binary labels. Received files, their hashes, and the author-reported source record are archived separately from the frozen evidence.

Table~\ref{tab:postfreeze-scorer} reports 96.0\% TSR agreement, with 176 true positives, 16 true negatives, eight false negatives, and no false positives. Agreement is 95.6\% on the 160 core runs, 100\% on the 20 repeated-poison runs, and 95.0\% on the 20 cross-model runs; by task family it is 95.5\% numerical and 96.7\% semantic/schema. Behavioral metrics use the 77 exposed trajectories. The scorer identifies all four human VPA cases, two each from Double-pass and Guard under repeated poisoning, providing additional examples of adoption after checking alongside the four historical adjudicated cases.

\begin{table}[t]
\centering\small
\setlength{\tabcolsep}{4pt}
\begin{tabular}{@{}lrrrrrr@{}}
\toprule
Metric & $n$ & Human $+$ & Precision & Recall & F1 & Agreement \\
\midrule
TSR & 200 & 184 & 1.000 & 0.957 & 0.978 & 0.960 \\
PAR & 77 & 14 & 1.000 & 0.857 & 0.923 & 0.974 \\
BCR & 77 & 10 & 1.000 & 0.800 & 0.889 & 0.974 \\
ADR & 77 & 1 & -- & 0.000 & 0.000 & 0.987 \\
VR & 77 & 64 & 1.000 & 0.953 & 0.976 & 0.961 \\
VPA & 77 & 4 & 1.000 & 1.000 & 1.000 & 1.000 \\
RR & 77 & 60 & 1.000 & 0.867 & 0.929 & 0.896 \\
\bottomrule
\end{tabular}

\caption{Frozen scorer versus the new human reference. TSR uses all 200 trajectories; behavior metrics use 77 exposed trajectories. Human $+$ denotes positive labels. ADR precision is undefined because the scorer predicts no positives.}
\label{tab:postfreeze-scorer}
\end{table}

\begin{figure}[t]
\centering
\includegraphics[width=\linewidth]{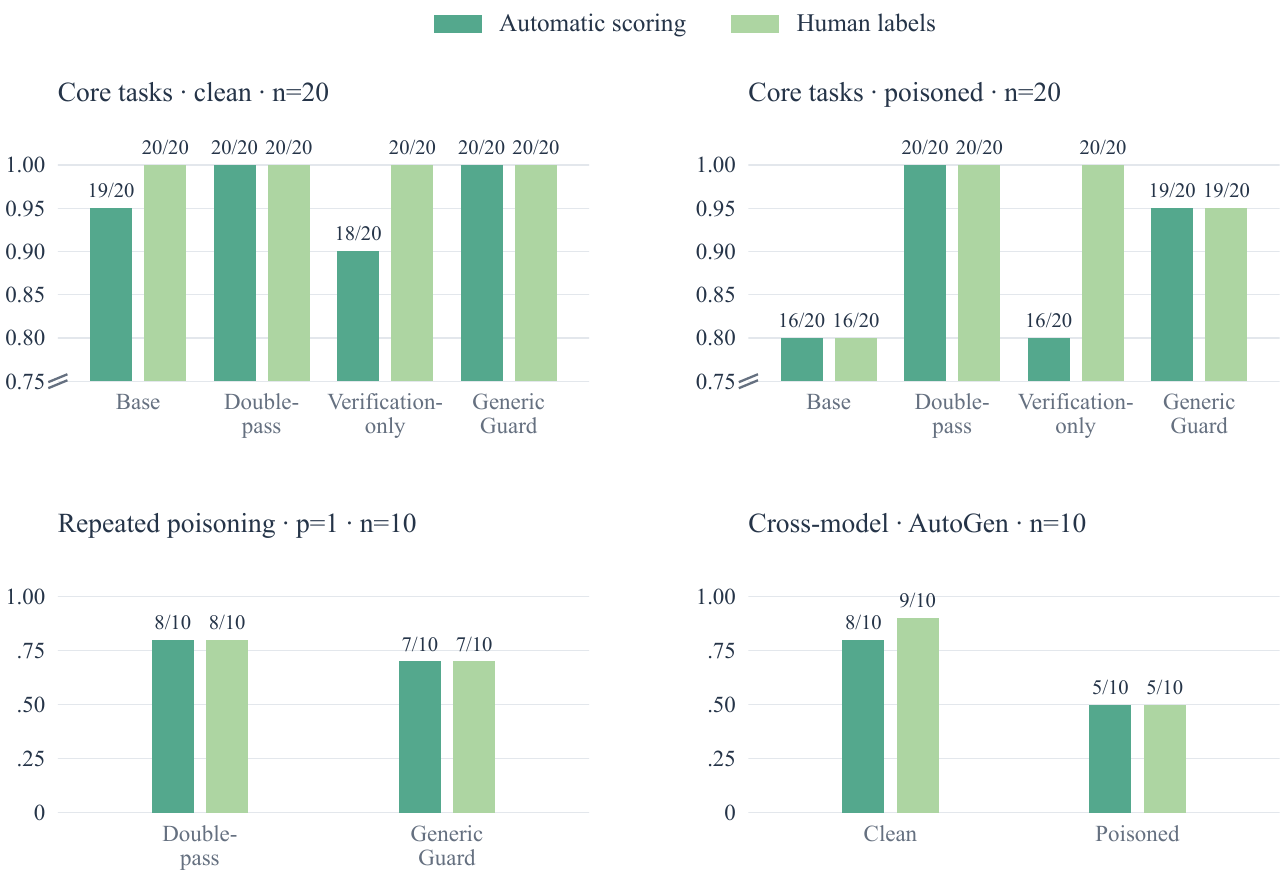}
\caption{Automatic and human task success on the same post-freeze trajectories. Labels show correct/total counts. Core conditions use 20 tasks and axes starting at 0.75; repeated-poison and cross-model conditions use ten tasks and zero-based axes. Only executed conditions are shown.}
\label{fig:postfreeze-methods}
\end{figure}

Figure~\ref{fig:postfreeze-methods} shows a human Double-pass--Base poisoned-TSR difference of $+0.20$ [0.05, 0.35], matching the automatic contrast. Human clean TSR is 1.00 for all four core methods, so its clean-adjusted contrast is also $+0.20$. Human Verification-only ties Double-pass at 1.00, compared with automatic TSR 0.80; Guard reaches 0.95 under both ratings. Repeated-poison TSR agrees across rating sources at 0.80 for Double-pass and 0.70 for Guard; their Guard--Double-pass contrast is $-0.10$ [$-0.30$, 0.00]. Intervals use 5,000 task-paired resamples within suites (seed 20260915) and describe the selected tasks, not population-wide method rankings.

The remaining TSR disagreements comprise six Verification-only cases, one Base case, and one AutoGen case; four of the Verification-only misses occur in poisoned runs. The seven-metric comparison contains 14 distinct disagreement trajectories, retained in the released error inventory. RR recall is 0.867 and BCR recall 0.800; ADR has only one human-positive exposed case and VPA four. These counts characterize metric-specific coverage alongside the overall agreement. The scorer remains fixed at the pre-annotation version (98c2ed8); none of these labels is used to tune the reported parser. The historical audits below use different samples and references and are reported separately.

The development packet contains 120 exposed trajectories, stratified equally across numerical, semantic/schema, and multi-table tasks. Two independent annotators saw queries, oracles, answers, and ordered events, but not scorer labels, model/adapter names, source paths, or each other's decisions. Twelve trajectories had disagreements in 16 label cells, resolved by separate blind adjudication. Table~\ref{tab:appendix-scorer-precision} distinguishes scorer accuracy from annotator agreement (macro agreement 0.967, $\kappa=0.913$). This packet did not separately label TSR, PAR, or VPA. A targeted review of 18 non-disputed VR-positive cases found all 18 relevant, but is not an independent accuracy estimate. The earlier 80-run author repeat-pass sheet is diagnostic only.

\begin{table}[t]
\centering
\small
\setlength{\tabcolsep}{5pt}
\begin{tabular}{@{}lrrrrrr@{}}
\toprule
Metric & $n$ & Precision & Recall & F1 & Agreement & $\kappa$ \\
\midrule
BCR & 120 & 0.750 & 0.250 & 0.375 & 0.983 & 0.914 \\
ADR & 120 & 1.000 & 0.030 & 0.059 & 0.925 & 0.826 \\
VR & 120 & 0.683 & 0.958 & 0.798 & 0.975 & 0.948 \\
RR & 120 & 0.840 & 0.840 & 0.840 & 0.983 & 0.965 \\
\bottomrule
\end{tabular}
\caption{Development audit on 120 trajectories: current scorer precision/recall/F1 against adjudicated labels, and original inter-annotator agreement before adjudication. These cases informed development; low BCR and ADR recall remains visible.}
\label{tab:appendix-scorer-precision}\label{tab:appendix-human-iaa}
\end{table}

A subsequent fixed packet contains 240 trajectories from 40 task IDs, excluding the 68 IDs used in development: 160 core cases from 20 tasks (ten per suite), four LangGraph methods, and clean/toxic pairs with Claude Haiku; 40 repeated-poison cases from Double-pass and Guard on those same tasks; and 40 AutoGen--Claude Sonnet clean/toxic cases on 20 other numerical tasks. It thus contains 100 environment pairs and 20 method pairs, with 94 exposed trajectories (56 core, 28 repeated, ten cross-model). Eligible task IDs were selected in sorted order, so the packet supports subset comparisons, not representative full-benchmark estimates. Rows were shuffled and explicit method/scorer fields masked, although pair slots and trace content could reveal protocols.

Two annotators labelled final correctness, poisoned-answer adoption, anomaly detection, substantive validation, recovery, and ambiguity. Table~\ref{tab:holdout-iaa} reports agreement before adjudication. A third annotator independently labelled all six fields on 60 disputed trajectories and ten additional shared-label ROAS cases selected to check the existing numeric tolerance; all ten were judged correct. These 70 full-row judgments replace prior labels, while 170 agreed rows retain shared labels. No final reference row is ambiguous.

\begin{table}[t]
\centering
\small
\begin{tabular}{@{}lrrrr@{}}
\toprule
Label & $n$ & Disagreements & Agreement & Cohen's $\kappa$ \\
\midrule
Final correctness & 240 & 27 & 0.888 & 0.718 \\
Poisoned-answer adoption & 240 & 3 & 0.988 & 0.721 \\
Anomaly detection & 240 & 34 & 0.858 & 0.091 \\
Substantive validation & 240 & 0 & 1.000 & 1.000 \\
Recovery after a check & 240 & 2 & 0.992 & 0.980 \\
Ambiguity & 240 & 0 & 1.000 & -- \\
\bottomrule
\end{tabular}
\caption{Agreement between the two returned holdout annotations. Neither annotator marked an ambiguous case, so its $\kappa$ is undefined. Final-correctness agreement is 0.944 on core cases, 0.925 under repeated poisoning, and 0.625 on cross-model cases.}
\label{tab:holdout-iaa}
\end{table}

Table~\ref{tab:holdout-scorer} places the original scorer and post-repair development recheck side by side against unchanged historical references and labels. The original scorer misses 64 of 193 human-correct answers and 36 of 72 recoveries (TSR agreement 173/240). The repair reaches 238/240 agreement, but these cases informed the repair and are no longer independent validation. Only three BCR and four VPA positives are available; high recheck recall therefore does not establish broad reliability.

\begin{table}[t]
\centering\small
\setlength{\tabcolsep}{4pt}
\begin{tabular}{@{}lrrrrrr@{}}
\toprule
& & & \multicolumn{2}{c}{Original scorer} & \multicolumn{2}{c}{Repair recheck} \\
Metric & $n$ & Human $+$ & Precision & Recall & Precision & Recall \\
\midrule
TSR & 240 & 193 & 0.977 & 0.668 & \best{1.000} & \best{0.990} \\
PAR & 94 & 7 & \best{1.000} & 0.714 & \best{1.000} & \best{1.000} \\
BCR & 94 & 3 & \best{1.000} & 0.333 & \best{1.000} & \best{1.000} \\
ADR & 94 & 4 & 0.667 & 0.500 & -- & -- \\
VR & 94 & 81 & \best{1.000} & \best{0.988} & \best{1.000} & \best{0.988} \\
VPA & 94 & 4 & \best{1.000} & \best{1.000} & \best{1.000} & \best{1.000} \\
RR & 94 & 72 & \best{1.000} & 0.500 & \best{1.000} & \best{0.972} \\
\bottomrule
\end{tabular}
\caption{Historical audit and development recheck using the same labels. Blue compares precision or recall between versions within each row, including ties. Original scorer: 4ae42cd; repaired parser: 98c2ed8. The recheck does not report ADR; full F1 values remain in released analyses. These are not independent estimates of revised-parser accuracy.}
\label{tab:holdout-scorer}\label{tab:development-recheck-v2}
\end{table}

\begin{table}[t]
\centering
\small
\begin{tabular}{@{}llrrr@{}}
\toprule
Setting & Method & Tasks & Automatic & Human \\
\midrule
One-shot & Base & 20 & \best{0.60} & \best{0.85} \\
 & Double-pass & 20 & \best{0.60} & \best{0.85} \\
 & Verification-only & 20 & 0.55 & \best{0.85} \\
 & Generic Guard & 20 & 0.50 & \best{0.85} \\
\midrule
Repeated, $p=1$ & Double-pass & 20 & \best{0.45} & \best{0.80} \\
 & Generic Guard & 20 & 0.25 & 0.70 \\
\bottomrule
\end{tabular}
\caption{Archived pre-repair automatic and human TSR on 20 shared audit tasks, with original references. Blue marks the highest TSR within each setting and rating source, including ties. The revised full analysis uses 118 retained instances and is reported separately.}
\label{tab:holdout-methods}
\end{table}

On the 20 core tasks, Guard minus Double-pass poisoned TSR is $-0.10$ [$-0.25$, $0.00$] automatically and $0.00$ [$0.00$, $0.00$] by humans; under repeated poisoning the differences are $-0.20$ [$-0.40$, $-0.05$] and $-0.10$ [$-0.25$, $0.00$]. Intervals use 5,000 paired task resamples within suites. The zero-width interval reflects identical observed labels, not population certainty. All four methods have human clean TSR 0.85 on the core subset; the separate cross-model subset has human TSR 0.75 clean and 0.60 toxic. Because both automatic and human core scores tie Base and Double-pass, this subset neither confirms nor refutes the full-matrix advantage. The four adjudicated VPA cases (one of 13 exposed Double-pass runs; three of 15 Guard runs) establish adoption after checking, not necessarily rejection of sufficient correct evidence.

\subsection{Frozen revision and reference audit}
\label{sec:revision-v2}
\label{sec:scorer-sensitivity}

The versioned comparison covers 5,396 archived trajectories. For each it stores the baseline decision (4ae42cd), frozen-parser decision (98c2ed8) with original references, and revised decision with corrected references. Raw answers, events, original task files, and human labels remain intact. Input hashes and selected claim spans accompany the versioned scoring analysis in the anonymized supplementary material. Table~\ref{tab:appendix-scorer-revision-impact} uses a different legacy baseline, f0aa054, and combines parser and reference changes on the same 944 retained defense trajectories; it must not be read as a parser-only effect.

\begin{table}[t]
\centering
\scriptsize
\setlength{\tabcolsep}{3pt}
\begin{tabular}{@{}lrrrrrrr@{}}
\toprule
Variant & Clean TSR & Poisoned TSR & BCR & PAR & VPA & VR & RR \\
\midrule
Base & 0.92$\to$0.99 & 0.83$\to$0.89 & 0.11$\to$0.09 & 0.27$\to$0.09 & 0.15$\to$0.00 & 0.46$\to$0.46 & 0.46$\to$0.44 \\
Double-pass & 0.92$\to$0.99 & 0.93$\to$1.00 & 0.01$\to$0.00 & 0.17$\to$0.00 & 0.16$\to$0.00 & 0.98$\to$0.98 & 0.97$\to$0.98 \\
Verification only & 0.93$\to$0.92 & 0.93$\to$0.90 & 0.02$\to$0.00 & 0.29$\to$0.01 & 0.27$\to$0.01 & 0.97$\to$0.97 & 0.96$\to$0.86 \\
Generic guard & 0.94$\to$0.98 & 0.94$\to$0.97 & 0.00$\to$0.00 & 0.24$\to$0.00 & 0.24$\to$0.00 & 0.97$\to$0.97 & 0.95$\to$0.94 \\
\bottomrule
\end{tabular}
\caption{Legacy $\to$ current scorer on identical defense trajectories. Behavior rates use the same exposed denominators as Table~\ref{tab:appendix-defense-exposure}.}
\label{tab:appendix-scorer-revision-impact}
\end{table}

Independent arithmetic over the 60 expanded numerical tasks found seven out-of-tolerance instances, two omitted ties, and two ambiguous channel-ranking queries. Corrections include conversion rate $615/3800=16.1842\%$, click-through rate $3800/49000=7.7551\%$, return rate $61/1100=5.5455\%$, units per order $1100/515=2.1359$, and three repeated instances. Both A100 and B200 attain revenue per unit 200 and are accepted. Two queries leave the conversion-rate denominator unspecified (clicks versus impressions), changing the winner; both are excluded across all revised methods. This is not exhaustive semantic-oracle validation.

The corrected analysis retains 118 single-table instances, 944 defense trajectories, and 1,572 stress trajectories. Historical poison targets and exposure events are unchanged. Two annotators also completed the separate review of 40 affected historical audit rows, agreeing on all six fields. Twelve trajectories have ambiguous, ineligible references and are excluded from accuracy comparisons; all 28 eligible trajectories are human-correct and match the frozen scorer. All seven exposed trajectories in this review belong to the ineligible group, leaving no eligible behavioral-rate comparison. These results document the reference corrections and are not pooled with the 200-case post-freeze evaluation. Original historical labels remain preserved alongside the new review.

\section{Agent Protocols and Framework Boundaries}
\label{sec:appendix-langgraph-replication}
\label{sec:appendix-framework-boundaries}

Base uses one ordinary LangGraph ReAct route. Double-pass runs two ordinary routes without handing the first answer to the second, and selects the second response verbatim. Verification-only instead passes the first answer as an explicitly untrusted claim and requests recomputation. Generic Guard additionally gives both the primary and verification routes a generic expectation prompt about ranges, bindings, and source evidence. It always selects the verification response, not an oracle-based choice between answers. Each route can involve multiple model requests: two routes do not mean two calls.

The three two-route variants share a ten-step cap per route, temperature zero, and a 3,072-output-token cap per model request. Equal caps do not imply equal realized tokens, calls, or latency. Both routes access the same dataframe and backend. Recomputing rows, reconstructing a join, or inspecting source evidence can establish VR; paraphrasing the first answer or producing a critique alone cannot. Passing an untrusted answer may still anchor the verifier. Always selecting the second response can correct poison or override a correct answer; clean TSR measures the net effect, but no independently adjudicated false-override audit is available.

LangGraph, smolagents, and AutoGen are the common observation-level adapters across GPT, Claude, and Qwen. The exact cross-model API identifiers are gpt-5.4-mini, claude-sonnet-4-6, and Qwen3.6-35B-A3B-no-thinking, respectively; the last is abbreviated as Qwen3.6-35B in the tables. DA-Agent completed the GPT numerical evaluation but not the other model families within the runtime budget. Its action/trace boundary differs and is reported separately. The historical PandasAI adapter modifies the final chat response only after the agent has finished; the agent cannot inspect that change. Those logs measure final-output integrity and remain archived, but are excluded from behavioral comparisons, cross-model means, and operator profiles.

The 13-task multi-table extension exposes auxiliary CSV tables alongside the primary dataframe. It covers table discovery and customer/order, support/team, clinic/region, and product/category joins. Verification reloads named tables and reconstructs joins from raw keys instead of copying a joined summary. This is a fresh execution route, not source diversity; it cannot detect shared CSV, parser, or join-key errors.

Additional six-step policies use abstention on route disagreement, a deterministic 0.5 task-level gate for recomputation, or keyword-based selective verification for rankings, ratios, denominators, schema, and retrieval. Table~\ref{tab:appendix-stronger-baselines} retains these fixed-subset diagnostics. Abstention lowers detected BCR but also answer coverage; the gates are not learned or generally calibrated.

\begin{table}[t]
\centering
\small
\setlength{\tabcolsep}{3.8pt}
\begin{tabular}{@{}llrrrrrr@{}}
\toprule
Suite & Adapter & Clean TSR & Poisoned TSR & BCR & VR & RR & PDR \\
\midrule
Numerical-20 & Abstain & 0.65 & 0.35 & \best{0.00} & 1.00 & 0.46 & 0.65 \\
Numerical-20 & Randomized & \best{0.95} & \best{0.80} & 0.23 & 0.62 & \best{0.54} & 0.65 \\
Numerical-20 & Selective & 0.90 & 0.70 & 0.25 & 0.50 & 0.42 & 0.60 \\
Semantic-10 & Abstain & 0.60 & 0.50 & \best{0.00} & 0.90 & 0.50 & 1.00 \\
Semantic-10 & Randomized & \best{1.00} & \best{0.80} & 0.20 & 0.70 & \best{0.70} & 1.00 \\
Semantic-10 & Selective & 0.90 & 0.70 & \best{0.00} & 1.00 & \best{0.70} & 1.00 \\
\bottomrule
\end{tabular}
\caption{LangGraph alternative verification policies on fixed stratified subsets. Blue marks higher TSR/RR and lower BCR within each suite, including ties. Behavior rates use poison-exposed runs; PDR gives the exposed fraction.}
\label{tab:appendix-stronger-baselines}
\end{table}

\section{Supplementary Results and Statistical Sensitivity}

\begin{figure*}[t]
\centering
\includegraphics[width=\textwidth]{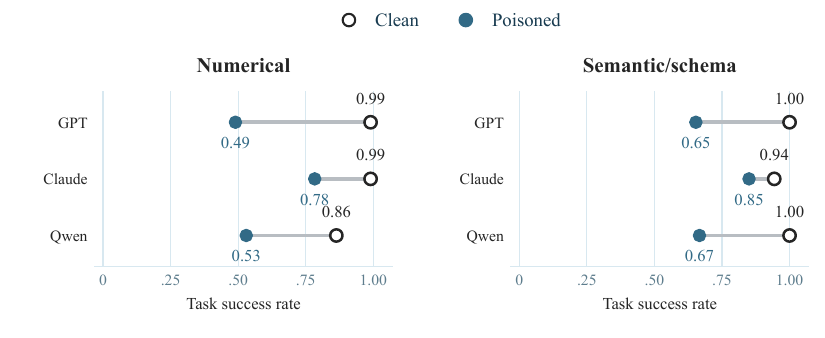}
\caption{Clean-to-poisoned task success by model and suite. Open circles denote clean TSR and filled circles poisoned TSR; connecting segments show their differences. Each point is an unweighted mean over LangGraph, smolagents, and AutoGen, not an individual-adapter result or a confidence interval. Table~\ref{tab:cross-model-rates} retains the numeric rates, BCR, and PDR.}
\label{fig:cross-agent-model}
\end{figure*}

Table~\ref{tab:appendix-guard-suite-breakdown} combines task-family results and exposure denominators. Conditional rates use different exposed subsets and are not direct rankings of trust calibration. Table~\ref{tab:operator-profile} gives the exact historical operator means for the main-text figure; sign flip is excluded because of off-target delivery. Figure~\ref{fig:cross-agent-model} and Table~\ref{tab:cross-model-rates} retain the cross-model summary and its numeric values.

\begin{table}[t]
\centering
\small
\setlength{\tabcolsep}{3pt}
\begin{tabular}{@{}llrrrrrrr@{}}
\toprule
Suite & Variant & Clean & Toxic & BCR & VR & RR & Exposed/$n$ & PDR \\
\midrule
Numerical & Base & \best{1.00} & 0.81 & 0.20 & 0.03 & 0.00 & 40/58 & 0.690 \\
Numerical & Double-pass & 0.98 & \best{1.00} & \best{0.00} & 0.95 & \best{0.95} & 41/58 & 0.707 \\
Numerical & Verification only & 0.95 & 0.90 & \best{0.00} & 0.95 & 0.83 & 41/58 & 0.707 \\
Numerical & Generic guard & 0.98 & 0.98 & \best{0.00} & 0.93 & 0.91 & 44/58 & 0.759 \\
Semantic & Base & 0.98 & 0.97 & 0.02 & 0.76 & 0.75 & 59/60 & 0.983 \\
Semantic & Double-pass & \best{1.00} & \best{1.00} & \best{0.00} & 1.00 & \best{1.00} & 60/60 & 1.000 \\
Semantic & Verification only & 0.90 & 0.90 & \best{0.00} & 0.98 & 0.88 & 60/60 & 1.000 \\
Semantic & Generic guard & 0.98 & 0.97 & \best{0.00} & 1.00 & 0.97 & 60/60 & 1.000 \\
\bottomrule
\end{tabular}
\caption{Suite-level TSR, exposure-conditioned behavior, and delivery (PDR). Blue marks higher TSR/RR and lower BCR within each suite, including ties. TSR uses all 58 numerical or 60 semantic instances. Detected BCR events are 8/40 for numerical Base, 1/59 for semantic Base, and zero for other variants.}
\label{tab:appendix-guard-suite-breakdown}\label{tab:appendix-defense-exposure}
\end{table}

\begin{table}[t]
\centering\small
\setlength{\tabcolsep}{5pt}
\begin{tabular}{@{}lrrrrr@{}}
\toprule
Operator & Poisoned TSR & BCR & VR & RR & PDR \\
\midrule
\multicolumn{6}{@{}l}{\emph{Numerical}} \\
Aggregate scale & 0.50 & 0.62 & 0.21 & 0.22 & 0.68 \\
Rank swap & 0.58 & 0.37 & 0.17 & 0.17 & 0.90 \\
\addlinespace[4pt]
\multicolumn{6}{@{}l}{\emph{Semantic/schema}} \\
Label swap & 0.59 & 0.55 & 0.21 & 0.21 & 0.74 \\
Treatment/control & 0.59 & 0.50 & 0.18 & 0.18 & 0.89 \\
Stale metadata & 0.65 & 0.33 & 0.67 & 0.65 & 1.00 \\
Column swap & 0.96 & 0.00 & 0.46 & 0.46 & 1.00 \\
Biased retrieval & 0.69 & 0.22 & 0.44 & 0.35 & 1.00 \\
\bottomrule
\end{tabular}

\caption{Full historical operator rates underlying Figure~\ref{fig:operator-profile}. BCR, VR, and RR means omit unexposed model--adapter groups; TSR and PDR include all groups. PDR measures delivery coverage, not agent robustness. Sign flip is excluded due to off-target delivery.}
\label{tab:operator-profile}
\end{table}

\begin{table}[t]
\centering\small
\setlength{\tabcolsep}{4pt}
\begin{tabular}{@{}llrrrr@{}}
\toprule
Suite & Model & Clean TSR & Poisoned TSR & BCR & PDR \\
\midrule
Numerical & GPT & \best{0.99} & 0.49 & 0.62 & 0.77 \\
 & Claude & \best{0.99} & \best{0.78} & \best{0.20} & 0.87 \\
 & Qwen & 0.86 & 0.53 & 0.40 & 0.73 \\
\midrule
Semantic/schema & GPT & \best{1.00} & 0.65 & 0.35 & 0.94 \\
 & Claude & 0.94 & \best{0.85} & \best{0.10} & 0.99 \\
 & Qwen & \best{1.00} & 0.67 & 0.35 & 0.93 \\
\bottomrule
\end{tabular}

\caption{Cross-model rates underlying Figure~\ref{fig:cross-agent-model}, averaged over the same three adapters. Blue marks higher TSR and lower BCR within each suite, including ties. BCR conditions on delivered poison within each adapter; PDR reports delivery coverage. These are means of adapter rates, not pooled trajectory rates.}
\label{tab:cross-model-rates}
\end{table}

\subsection{Paired inference and measured cost}
\label{sec:appendix-bootstrap-ci}
\label{sec:appendix-latency-distribution}

For Guard and Double-pass success $G_{i,e}$ and $D_{i,e}$ on task $i$ in environment $e$, the clean-adjusted contrast is
\[
\Delta_{\mathrm{interaction}}=\frac{1}{n}\sum_i
[(G_{i,\mathrm{toxic}}-D_{i,\mathrm{toxic}})
 -(G_{i,\mathrm{clean}}-D_{i,\mathrm{clean}})].
\]
Table~\ref{tab:appendix-paired-defense-ci} preserves all four outcomes in each task resample (5,000 draws, seed 13), with 58 numerical and 60 semantic instances in combined draws. Its combined interval includes zero. Behavior contrasts require exposure in both variants and need not equal differences of marginal rates. Zero detected BCR differences do not prove equal human-judged risk.

\begin{table}[t]
\centering
\scriptsize
\setlength{\tabcolsep}{3pt}
\begin{tabular}{@{}llrrr@{}}
\toprule
Suite & Metric & $n_{\mathrm{paired}}$ & $\Delta$ (Guard$-$DP) & 95\% CI \\
\midrule
Numerical & Clean TSR & 58 & +0.000 & [-0.052, 0.052] \\
Numerical & Poisoned TSR & 58 & -0.017 & [-0.052, 0.000] \\
Numerical & $\Delta_{\mathrm{interaction}}$ & 58 & -0.017 & [-0.086, 0.035] \\
Numerical & BCR & 41 & +0.000 & [0.000, 0.000] \\
Numerical & VR & 41 & +0.000 & [-0.073, 0.073] \\
Numerical & RR & 41 & -0.024 & [-0.122, 0.049] \\
Semantic/schema & Clean TSR & 60 & -0.017 & [-0.050, 0.000] \\
Semantic/schema & Poisoned TSR & 60 & -0.033 & [-0.083, 0.000] \\
Semantic/schema & $\Delta_{\mathrm{interaction}}$ & 60 & -0.017 & [-0.083, 0.033] \\
Semantic/schema & BCR & 60 & +0.000 & [0.000, 0.000] \\
Semantic/schema & VR & 60 & +0.000 & [0.000, 0.000] \\
Semantic/schema & RR & 60 & -0.033 & [-0.083, 0.000] \\
Combined & Clean TSR & 118 & -0.009 & [-0.034, 0.017] \\
Combined & Poisoned TSR & 118 & -0.025 & [-0.059, 0.000] \\
Combined & $\Delta_{\mathrm{interaction}}$ & 118 & -0.017 & [-0.059, 0.017] \\
\bottomrule
\end{tabular}
\caption{Task-paired bootstrap differences between Generic Guard and the matched Double-pass (DP). TSR and interaction use all shared tasks; BCR, VR, and RR use tasks exposed in both variants. Bootstrap uses 5,000 task resamples and seed 13; combined TSR resampling is stratified by suite. Degenerate BCR difference intervals describe the observed scorer labels, not proof of equal human-judged risk.}
\label{tab:appendix-paired-defense-ci}
\end{table}

The 118 retained instances form 87 template clusters across 30 CSVs. Templates group dataset, query with generic checking suffixes removed, and poisoning specification without severity/probability bookkeeping. Dataset resampling assigns shared sources the same multiplicity across suites; combined estimates preserve suite weights (5,000 draws, seed 20260916). Table~\ref{tab:cluster-sensitivity-v2} assesses this dependence, not scorer error or API sampling variance. Released equal-cluster-weight means answer a different question from instance-weighted means.

\begin{table}[t]
\centering\small
\begin{tabular}{@{}lrrr@{}}
\toprule
Poisoned TSR comparison & Difference & Template 95\% CI & Dataset 95\% CI \\
\midrule
Double-pass $-$ Base & +0.110 & [0.056, 0.171] & [0.060, 0.172] \\
Guard $-$ Double-pass & -0.025 & [-0.054, 0.000] & [-0.056, 0.000] \\
Verification $-$ Double-pass & -0.102 & [-0.163, -0.049] & [-0.165, -0.047] \\
\bottomrule
\end{tabular}
\caption{Paired sensitivity to template and dataset dependence, with corrected references and the frozen parser. All comparisons use the same retained task instances.}
\label{tab:cluster-sensitivity-v2}
\end{table}

Marginal task-bootstrap intervals are provided in the supplementary statistical results (5,000 draws, seed 13 plus adapter index). For zero-event BCR cells, exact one-sided 95\% binomial upper bounds on detected-event rates are approximately 0.07 for the numerical two-route variants and 0.05 for semantic variants; a degenerate empirical bootstrap is not evidence of zero true risk.

Table~\ref{tab:appendix-latency-distribution} retains timing over the original 120 tasks, including two subsequently excluded queries; the main-text means use the retained 118. Tool-event counts are measured costs, not token billing, which the gateway does not provide. For the 13 join tasks at $p=1$, mean latency is 41.59 seconds for Verification-only, 61.48 for Double-pass, and 54.34 for Guard; these small-sample timings are sensitive to backend queueing.

\begin{table}[t]
\centering
\small
\setlength{\tabcolsep}{4pt}
\begin{tabular}{@{}lrrrrr@{}}
\toprule
Suite and adapter & Rows & Mean sec. & P50 sec. & P90 sec. & Tool events \\
\midrule
Numerical Base & 60 & \best{16.35} & \best{14.82} & \best{27.37} & \best{2.32} \\
Numerical Double-pass & 60 & 31.22 & 25.65 & 50.31 & 4.45 \\
Numerical Verification only & 60 & 31.69 & 29.61 & 52.93 & 4.30 \\
Numerical Generic guard & 60 & 40.19 & 39.46 & 60.35 & 4.62 \\
Semantic/schema Base & 60 & \best{17.60} & \best{16.33} & \best{30.89} & \best{2.40} \\
Semantic/schema Double-pass & 60 & 41.10 & 37.97 & 65.86 & 4.77 \\
Semantic/schema Verification only & 60 & 36.34 & 31.99 & 55.60 & 4.57 \\
Semantic/schema Generic guard & 60 & 39.61 & 36.65 & 57.13 & 5.10 \\
\bottomrule
\end{tabular}
\caption{Historical toxic-run latency and event counts on all 120 released instances, including two reference-excluded queries. Blue marks lower measured cost within each suite, not overall method quality.}
\label{tab:appendix-latency-distribution}
\end{table}

\subsection{Repeated poisoning and evidence availability}

Repeated poisoning removes the shared one-shot constraint and applies a deterministic hash gate with $p\in\{0.25,0.50,0.75,1.00\}$ to eligible operator-matching returns. The original matrix has 133 tasks, three two-route variants, and four probabilities (1,596 trajectories); reference exclusions retain 1,572. There is no single-route Base control. Even at $p=1$, nonmatching calls can return unmodified evidence. This stresses returned observations, not a compromised database.

\begin{table}[t]
\centering
\small
\setlength{\tabcolsep}{6pt}
\resizebox{\linewidth}{!}{%
\begin{tabular}{@{}llrrrrrrr@{}}
\toprule
Suite & Variant & Poisoned TSR & BCR & PAR & VPA & VR & RR & PDR \\
\midrule
Numerical & Double-pass & \best{0.81} & \best{0.00} & 0.20 & 0.20 & 1.00 & \best{0.72} & 0.69 \\
Numerical & Verification only & 0.71 & \best{0.00} & 0.27 & 0.27 & 1.00 & 0.59 & 0.71 \\
Numerical & Generic guard & 0.79 & \best{0.00} & \best{0.11} & \best{0.11} & 0.98 & 0.70 & 0.76 \\
Semantic/schema & Double-pass & \best{0.95} & \best{0.00} & 0.05 & 0.05 & 0.90 & \best{0.85} & 1.00 \\
Semantic/schema & Verification only & 0.87 & \best{0.00} & \best{0.03} & \best{0.03} & 0.95 & 0.82 & 1.00 \\
Semantic/schema & Generic guard & 0.82 & \best{0.00} & 0.12 & 0.12 & 1.00 & 0.82 & 1.00 \\
Multi-table & Double-pass & \best{0.54} & \best{0.00} & 0.38 & 0.38 & 1.00 & \best{0.54} & 1.00 \\
Multi-table & Verification only & \best{0.54} & \best{0.00} & 0.38 & 0.38 & 1.00 & \best{0.54} & 1.00 \\
Multi-table & Generic guard & 0.46 & \best{0.00} & \best{0.08} & \best{0.08} & 1.00 & 0.46 & 1.00 \\
\bottomrule
\end{tabular}%
}

\caption{Full-suite repeated poisoning at \(p=1\). Blue marks higher TSR/RR and lower BCR/PAR/VPA within each suite, including ties. Behavioral rates condition on exposed runs; PDR reports exposure.}
\label{tab:appendix-multiroute-stress}
\end{table}

\begin{figure*}[t]
\centering
\includegraphics[width=0.86\textwidth]{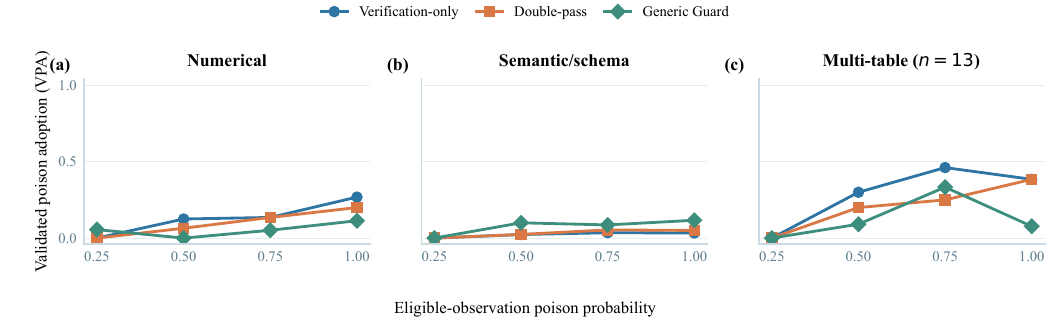}
\caption{Exposure-conditioned validated poison adoption (VPA) under repeated poisoning for the numerical, semantic/schema, and 13-task multi-table suites. The $p=1$ task-success and behavior rates are reported in Table~\ref{tab:appendix-multiroute-stress} and the released CSV. No method dominates across suites, and the small multi-table sample should not be read as a precise population estimate. The corruption acts on returned observations, not the shared source.}
\label{fig:guard-tradeoff}
\end{figure*}

At $p=1$, all nine cells have zero detected BCR but nonzero PAR and VPA. Some lower-probability cells have nonzero BCR; the supplementary results report these cells. The 13-task join estimates are imprecise and finite-sample curves need not be monotonic. In the revised inventory, 78 of 101 VPA trajectories have only corrupted qualifying checks, while 23 have at least one unmodified check. Only one has an automatically identifiable clean-reference match, still requiring human review. These are trace categories, not evidence that 101 agents rejected sufficient correct evidence. The four adjudicated VPA cases establish adoption after checking; a corrupted check remains a possible explanation.

The supplementary case studies illustrate why the metrics differ: fluent arithmetic can repeat a wrong entity binding; verbal doubt can leave the poisoned answer unchanged; a fresh group-by or reconstructed join can restore the clean answer. Conversely, additional checking can exhaust the budget or return another incorrect value. Thus reduced BCR alone does not establish recovery.

\section{Delivery Audit and Controlled Evidence Experiments}

\subsection{Historical delivery and repaired scalar controls}
\label{sec:delivery-v3}

We audit the observation logs for eight observation-level adapter identifiers in the primary, defense, and repeated-poison manifests. Of 3,319 recorded corruptions, 204 events in 157 trajectories are flagged because the legacy sign-flip operator selected the first number rather than the specified answer value. Overlapping subsets include 73 partial numeric tokens, 20 changes with no numerical or textual content difference after complete-number normalization (such as $0\rightarrow-0.0$), and six tool-error outputs. These are mechanically identifiable flags, not an exhaustive semantic validity assessment; unflagged events are not certified valid. Flags occur in 43 cross-model, 14 expanded-GPT, 13 AutoGen-defense, 21 LangGraph-defense, and 113 repeated-poison events.

The release preserves the original logs and reports behavioral rates both with recorded exposure and after excluding flagged events. This sensitivity holds final answers fixed: it cannot restore a consumed one-shot opportunity, change later observations, or predict how an agent would respond to a repaired intervention. TSR is therefore unchanged by event relabelling. As a separate common-task sensitivity, removing all eight sign-flip instances leaves 110 defense tasks. Double-pass minus Base TSR is $+0.118$ (task-paired 95\% bootstrap interval [0.064, 0.182]); Guard minus Double-pass is $-0.027$ [$-0.064$, 0.000], and Verification-only minus Double-pass is $-0.109$ [$-0.173$, $-0.055$]. This preserves the historical comparison on the remaining tasks, not validity of the excluded operator.

In the repeated-poison matrix, excluding flagged events reduces exposed trajectories from 1,054 to 997 while retaining all 101 VPA classifications. Table~\ref{tab:delivery-sensitivity-v3} reports the main GPT sensitivity.

\begin{table}[t]
\centering\small
\begin{tabular}{@{}lrrrr@{}}
\toprule
& \multicolumn{2}{c}{Recorded delivery} & \multicolumn{2}{c}{Flag-excluded sensitivity} \\
Adapter & $n_{\rm exposed}$ & BCR & $n_{\rm exposed}$ & BCR \\
\midrule
LangGraph ReAct & 101 & 0.376 & 99 & 0.384 \\
smolagents & 87 & 0.529 & 83 & 0.554 \\
AutoGen & 101 & 0.267 & 93 & 0.290 \\
\bottomrule
\end{tabular}
\caption{Exposure sensitivity on the 118 retained expanded-GPT tasks. Answers and TSR are fixed; unflagged delivery is not exhaustively certified.}
\label{tab:delivery-sensitivity-v3}
\end{table}

The new injector rejects error outputs, parses complete numeric tokens including thousands separators and scientific notation, and targets the configured clean value. Numeric replacement requires a unique matching scalar or labelled-scalar line, excluding dataframe indices and multi-value lines; an optional explicit field name further restricts eligibility. Missing or ambiguous targets and value-preserving transformations neither count as exposure nor consume the one-shot opportunity. Rank swaps no longer replace an entire nonmatching observation with an oracle string. These conservative rules reduce false delivery but can also reduce delivery rates. A scalar match is not a proof of semantic field identity, so original and returned observations remain available for inspection. The answer parser, scoring tolerances, historical tasks, and frozen human packet are unchanged.

Before new answers were collected, we froze 16 development tasks: all eight original sign-flip instances, three other numerical tasks, and five semantic/schema tasks. Their references are unchanged by the arithmetic audit. Claude Haiku 4.5 runs Base in clean and toxic environments and three two-route methods under two toxic conditions: a shared one-shot opportunity or a fresh one-shot opportunity at each route boundary. This yields 128 trajectories. Methods share a ten-step cap per route and a 3,072-output-token cap per request, at temperature zero. Conditions are separate executions paired by task, not replay of an identical first route. Resetting eligibility does not force corruption and does not eliminate subsequent clean calls within the same route. Actual exposure by route is therefore reported alongside all-task TSR. The subset deliberately covers a discovered defect; it is not a representative prevalence estimate or an independent scorer test.

We additionally freeze six numerical tasks over the 344 observational rows of the CC0 Palmer Penguins dataset \citep{horst2020palmerpenguins}. Tasks cover means, filters, a category count, and unit conversion, with explicit missing-value handling and dataframe-computed references. Base clean/toxic and Double-pass under the two exposure conditions give 24 trajectories. The data, license, source revision, and hashes are released. This tests operation on a larger real table, not joins, independent datasets, unseen training data, or production deployment. All six tasks share one source, precluding a dataset-level generalization estimate.

All four public-data conditions achieve TSR $6/6$. Actual exposure occurs in $4/6$ Base toxic runs, $4/6$ shared Double-pass runs, and $5/6$ per-route Double-pass runs. Importantly, all four exposed Base returns retain an unchanged correctly rounded answer alongside the altered scalar: the operator changes one eligible occurrence, not every redundant expression of the answer. Thus this is a partial-corruption execution check, not evidence of recovery after correct evidence was removed, nor a demonstration of deployment robustness. Non-delivery and residual clean-reference candidates are released explicitly.

Tables~\ref{tab:fresh-control-v3} and~\ref{tab:fresh-public-v3} report all 152 completed trajectories. On the control subset, per-route minus shared TSR is $-0.0625$ for both Double-pass and Verification-only (paired 95\% intervals [$-0.1875$, 0.000]), and $-0.1875$ for Guard [$-0.375$, 0.000]. All intervals include zero. Base already solves $16/16$ toxic tasks, so this selected subset does not replicate the historical full-matrix retry advantage. The manipulation increases second-route exposure, but the outcome evidence is too small and selective to attribute that historical advantage to evidence availability alone. Partial corruption can also leave correct values within the same returned observation, not just in later calls.

\begin{table}[t]
\centering\small
\begin{tabular}{@{}llrrrr@{}}
\toprule
Method & Condition & $n$ & TSR & Exposed & Route 2 exposed \\
\midrule
Base & Clean & 16 & 1.000 & 0 & -- \\
Base & Shared & 16 & 1.000 & 13 & -- \\
Double-pass & Shared & 16 & 1.000 & 12 & 1 \\
Double-pass & Per-route & 16 & 0.938 & 12 & 12 \\
Verification & Shared & 16 & 0.875 & 13 & 1 \\
Verification & Per-route & 16 & 0.812 & 13 & 12 \\
Guard & Shared & 16 & 1.000 & 13 & 1 \\
Guard & Per-route & 16 & 0.812 & 14 & 10 \\
\bottomrule
\end{tabular}
\caption{Preselected 16-task control, using the repaired injector and frozen scorer. Exposed counts are trajectories with actual corruption, not opportunities. TSR uses all tasks.}
\label{tab:fresh-control-v3}
\end{table}
\begin{table}[t]
\centering\small
\begin{tabular}{@{}llrrrr@{}}
\toprule
Method & Condition & $n$ & TSR & Exposed & Route 2 exposed \\
\midrule
Base & Clean & 6 & 1.000 & 0 & -- \\
Base & Shared & 6 & 1.000 & 4 & -- \\
Double-pass & Shared & 6 & 1.000 & 4 & 0 \\
Double-pass & Per-route & 6 & 1.000 & 5 & 3 \\
\bottomrule
\end{tabular}
\caption{Six-task public-data transfer check, using the repaired injector and frozen scorer. Exposed counts are trajectories with actual corruption, not opportunities. TSR uses all tasks.}
\label{tab:fresh-public-v3}
\end{table}

The pre-execution protocol, included with the repaired-injector experiments in the anonymized supplementary material, fixes the tasks, methods, budgets, input hashes, and analyses. Per-job logs retain failures without outcome-based reruns. An HTTP-429 recovery amendment lowers concurrency and reruns only environments without a completed trajectory; completed answers are retained regardless of score. Historical event flags and exposure sensitivity are archived separately from fresh trajectories. Paired intervals use 5,000 task resamples with seed 20260916; they describe this selected task set and single execution per condition, not model-sampling uncertainty. Request counts denote logical model calls, excluding transport retries. Completed-trajectory latency excludes abandoned attempts, and mixed scheduling further limits cost comparisons. These measurements do not establish token-cost equivalence.

\subsection{Field-bound, matched-parent evidence controls}
\label{sec:evidence-v4}

The V3 public-table check leaves correct aliases in corrupted observations, and its route comparisons do not share a fixed first-stage transcript. We therefore introduce a separate structured-evidence experiment. It is not a rerun of the historical framework matrix or a validation of the free-text scorer. All conditions use the same JSON tool protocol and a general warning that outputs may be inaccurate. The experiment fixes an initial aggregate call rather than measuring whether an agent spontaneously selects or verifies that call.

An aggregate query is identified by its operation (mean, sum, or count), column, equality filters, and scaling factor. Its result contains a numerical value and two redundant four-decimal strings: a decimal representation and a comma-formatted display. Clean delivery leaves all three unchanged. Partial corruption alters only the numerical value; full-target corruption regenerates all three from the same incorrect scalar. A change must alter every alias to qualify as full-target delivery. Errors, value-preserving edits, and queries for other fields are not poisoned. We do not search arbitrary output text for numbers matching an oracle. Source files, raw-row queries, and alternative aggregates remain unchanged: full-target corruption does not mean the removal of all possible correct evidence. For example, an agent can check a corrupted mean using a sum and a count.

We freeze eight tasks each on Palmer Penguins (344 records), Auto MPG (398), and the hourly Bike Sharing table (17,379) \citep{horst2020palmerpenguins,quinlan1993autompg,fanaee2013bikedata}. The two UCI exports are distributed under CC BY 4.0; Penguins is CC0. Released source URLs, licenses, and file hashes accompany the data. Tasks cover means, filters, counts, scale conversions, and sums, with explicit missing-value rules. Standard-library and pandas implementations independently compute matching references. Incorrect scalars alternate between $1.25$ and $0.75$ times the reference, rounded to an integer for counts, before any model outcomes are observed. These are controlled synthetic interventions over real observational tables, not a sample of naturally occurring errors. These public datasets may have appeared in model training.

The 24 tasks each receive clean, partial, and full primary conditions (72 trajectories). One completed full-poison primary transcript per task is retained as the parent. In the second stage, \emph{verify} branches copy this entire transcript and ask for review; \emph{retry} branches receive the original question without the parent answer or history. Both policies receive a mandatory fresh target-aggregate observation under either clean or full corruption. Subsequent matching calls follow that condition persistently, while alternative queries remain unmodified. Each policy--condition combination has two completions per task (192 second-stage trajectories). Clean and full verify branches therefore share exactly the same earlier answer and tool history; conditions within retry share the same history-free task prompt. Branch execution order is randomized before outcomes. The policies are diagnostic controls, not replacements for the historical Double-pass or Generic Guard implementations.

Claude Haiku 4.5 runs at temperature zero with at most six model requests per stage and 1,536 output tokens per request, initially at concurrency two. The seeded aggregate call does not consume a model request. The final JSON numeric field is compared with the reference using absolute tolerance 0.01; explanations cannot override this field. Format errors receive a corrective prompt within the same budget; an exhausted trajectory without a valid final answer fails. This structured endpoint avoids free-text candidate selection but does not independently validate the historical scorer. Completed answers are never retried based on correctness. Exhausted API failures are archived, and per-request checkpoints preserve the completed prefix when a job resumes. Logical request counts exclude internal HTTP retries; token billing is unavailable.

Pre-specified contrasts are full-minus-clean and full-minus-partial primary correctness, and clean-minus-full second-stage correctness for each policy, with poisoned-answer adoption reported separately. The two completions are averaged within task before 5,000 paired bootstrap resamples stratified within the three fixed datasets. We report source-specific effects, actual delivery, unchanged alternative calls, and unresolved outcomes. Intervals condition on the selected sources, fixed parents, and limited completions; they do not quantify dataset-population generalization or comprehensive model-sampling uncertainty. Related queries share data, and no multiplicity-adjusted significance claim is made. Even a strong local evidence-condition effect would not identify what fraction of the historical retry gain it explains.

\begin{figure}[t]
\centering
\includegraphics[width=0.85\linewidth]{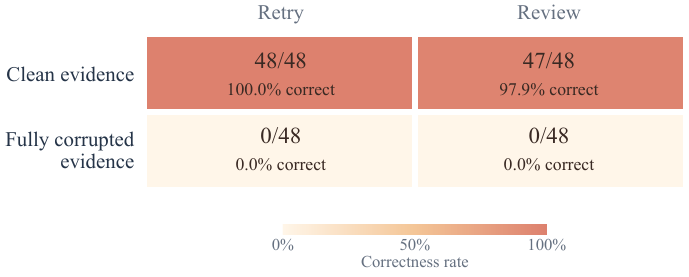}
\caption{Second-stage correctness under clean versus fully corrupted target evidence. Counts aggregate 24 tasks with two completions per policy--condition, not 48 independent tasks. Review branches inherit the fixed poisoned parent; retries restart without its history. The full 0--100\% color scale includes zero outcomes. Table~\ref{tab:evidence-v4} retains source-specific and first-stage counts.}
\label{fig:evidence-control}
\end{figure}

\begin{table}[t]
\centering\small
\begin{tabular}{@{}llrrr@{}}
\toprule
Stage/policy & Evidence & Penguins & Auto MPG & Bike \\
\midrule
Primary & clean & 8/8 & 8/8 & 8/8 \\
Primary & partial & 8/8 & 7/8 & 5/8 \\
Primary & full & 0/8 & 0/8 & 0/8 \\
Retry & clean & 16/16 & 16/16 & 16/16 \\
Retry & full & 0/16 & 0/16 & 0/16 \\
Verify & clean & 16/16 & 16/16 & 15/16 \\
Verify & full & 0/16 & 0/16 & 0/16 \\
\bottomrule
\end{tabular}
\caption{Structured-evidence V4 correctness counts on eight tasks per source. Second-stage cells contain two completions per task sharing a fixed full-poison primary. These are not 16 independent task instances. Full corruption replaces all three target-field aliases; alternative queries and raw rows remain unmodified.}
\label{tab:evidence-v4}
\end{table}

All 264 scheduled trajectories are retained, including one review that exhausts its budget without a final answer. One exhausted API failure is successfully resumed; 311 logical requests complete. The delivery audit finds 120 full-target and 24 partial-target corruptions. Every full-target event changes all three result aliases, with no unchanged clean alias in that field. First-stage correctness is $24/24$ clean, $20/24$ partial, and $0/24$ full; all 24 full-poison primary answers adopt the incorrect scalar. Full-minus-partial TSR is $-0.833$ (task-paired 95\% interval [$-0.958$, $-0.708$]). Thus partial corruption can substantially overstate resistance in this interface; it does not measure the magnitude of bias in the historical matrix.

Second-stage correctness is $48/48$ for clean retries and $47/48$ for clean reviews, versus $0/48$ for each policy under full corruption. The clean-minus-full contrast is $+1.000$ for retry and $+0.979$ [0.938, 1.000] for review. All-success/all-failure contrasts yield degenerate empirical bootstrap intervals, not certainty beyond the selected tasks. The only clean-review failure requests successive raw-value pages until the six-request budget expires; it does not adopt the poisoned answer. Under full corruption, all 48 retry answers and 46 of 48 review answers match the poisoned scalar. The other two reviews apply the requested scale a second time to the poisoned value, producing different incorrect numbers. Thus absence of a direct poisoned-value match need not mean rejection of its evidence.

The clean-review branches execute unmodified alternative queries in $10/48$ trajectories; no other condition makes a successful alternative query. In particular, full-poison reviews do not request additional checks beyond the mandated new observation. This experiment therefore supports a controlled effect of the supplied evidence and sensitivity to residual aliases. It does not measure whether agents choose to verify on their own or provide new human-confirmed VPA cases. It does not establish an advantage of verification instructions over retry, or identify how much of the historical Double-pass gain is caused by later clean evidence.

The pre-execution protocol for the structured-evidence controls, provided in the anonymized supplementary material, fixes the task list, order, input hashes, budget, and analysis endpoints before answers are collected. The completion audit checks every changed alias, source hashes, common-parent hashes, all scheduled conditions, and absence of duplicate completed jobs. The completed 200-trajectory human evaluation and 40-case reference review retain their pre-V4 evidence and are analyzed separately in Appendix~\ref{sec:postfreeze-human}; the structured experiment evaluates evidence conditions rather than free-text scoring.

\end{document}